\documentclass{article} 
\usepackage{iclr2027_conference,times}

\usepackage{amsmath,amsfonts,bm}

\def\eqref#1{equation~\ref{#1}}

\def\1{\bm{1}}

\DeclareMathAlphabet{\mathsfit}{\encodingdefault}{\sfdefault}{m}{sl}
\SetMathAlphabet{\mathsfit}{bold}{\encodingdefault}{\sfdefault}{bx}{n}

\usepackage{hyperref}
\usepackage{url}
\usepackage{xcolor}         

\usepackage[dvipsnames,svgnames,x11names]{xcolor}
\usepackage{tikz}
\usetikzlibrary{arrows.meta,shapes,calc,matrix,fit,positioning,backgrounds,decorations.markings}
\usepackage{pgfplots}
\usepackage{pgfplotstable}
\pgfplotsset{compat=1.9}
\usepackage{xstring}

\usepgfplotslibrary{external}

\IfBeginWith*{\jobname}{fig/extern/}{\finalcopy}{}

\tikzstyle{every picture}+=[
	remember picture,
	every text node part/.style={align=center},
	every matrix/.append style={ampersand replacement=\&},
]
\tikzstyle{tight} = [inner sep=0pt,outer sep=0pt]
\tikzstyle{node}  = [draw,circle,tight,minimum size=12pt,anchor=center]
\tikzstyle{op}    = [draw,circle,tight]
\tikzstyle{dot}   = [fill,draw,circle,inner sep=1pt,outer sep=0]
\tikzstyle{pt}    = [fill,draw,circle,inner sep=1.5pt,outer sep=.2pt]
\tikzstyle{box}   = [draw,rectangle,inner sep=3pt]
\tikzstyle{high}  = [black!60]
\tikzstyle{group} = [high,box,opacity=.5]
\tikzstyle{dim1}  = [fill opacity=.3,text opacity=1]
\tikzstyle{dim2}  = [fill opacity=.5,text opacity=1]
\tikzstyle{dim3}  = [fill opacity=.7,text opacity=1]
\tikzstyle{rectc} = [tight,transform shape]
\tikzstyle{rect}  = [rectc,anchor=south west]

\tikzset{every mark/.append style={solid}}
\pgfplotsset{
	grid=both, width=\columnwidth, try min ticks=5,
	every axis/.append style={font=\small},
	every axis plot/.append style={thick,mark=none,mark size=1.8,tension=0.18},
	legend cell align=left, legend style={fill opacity=0.8},
	xticklabel={\pgfmathprintnumber[assume math mode=true]{\tick}},
	yticklabel={\pgfmathprintnumber[assume math mode=true]{\tick}},
	nodes near coords math/.style={
		nodes near coords={\pgfmathprintnumber[assume math mode=true]{\pgfplotspointmeta}},
	},
}

\pgfplotsset{
	dash/.style={mark=o,dashed,opacity=0.6},
	dott/.style={mark=o,dotted,opacity=0.6},
	nolim/.style={enlargelimits=false},
	plain/.style={every axis plot/.append style={},nolim,grid=none},
}

\pgfdeclarelayer{bg4}
\pgfdeclarelayer{bg3}
\pgfdeclarelayer{bg2}
\pgfdeclarelayer{bg1}
\pgfdeclarelayer{fg1}
\pgfdeclarelayer{fg2}
\pgfdeclarelayer{fg3}
\pgfdeclarelayer{fg4}
\pgfsetlayers{bg4,bg3,bg2,bg1,main,fg1,fg2,fg3,fg4}

\pgfkeys{/tikz/.cd, aspect/.store in=\aspect, aspect=1}
\pgfkeys{/tikz/.cd, depth/.store in=\depth, depth=.5}
\pgfkeys{/tikz/.cd, stepx/.store in=\step, stepx=1}

\tikzstyle{geom} = [line join=bevel,aspect=1,depth=.5,z={(\depth*\aspect,\depth)}]
\tikzstyle{wire} = [geom,draw,thick]

\def\cx[#1,#2,#3]{#1}
\def\cy[#1,#2,#3]{#2}
\def\cz[#1,#2,#3]{#3}
\def\ex[#1,#2,#3]{#1,0,0}
\def\ey[#1,#2,#3]{0,#2,0}
\def\ez[#1,#2,#3]{0,0,#3}

\usepackage[export]{adjustbox}

\newcommand{\Th}[1]{\textsc{#1}}

\newcommand{\red}[1]{{\textcolor{red}{#1}}}

\newcommand{\citeme}[1]{\red{[XX]}}
\newcommand{\refme}[1]{\red{(XX)}}

\makeatletter
\newcommand*\bdot{\mathpalette\bdot@{.7}}
\newcommand*\bdot@[2]{\mathbin{\vcenter{\hbox{\scalebox{#2}{$\m@th#1\bullet$}}}}}
\makeatother

\makeatletter
\DeclareRobustCommand\onedot{\futurelet\@let@token\@onedot}
\def\@onedot{\ifx\@let@token.\else.\null\fi\xspace}

\makeatother

\newcommand{\ours}{\texttt{Latent-Foresight}\xspace}
\newcommand{\oursplus}{\texttt{Latent-Foresight$^+$}\xspace}

\definecolor{ForestGreen}{RGB}{34,139,34}
\definecolor{Orange}{RGB}{1.0, 0.55, 0.0}
\definecolor{Purple}{RGB}{0.58, 0.44, 0.86}
\definecolor{VibrantMagenta}{RGB}{255,0,197}
\definecolor{DarkenedMagenta}{RGB}{225,0,167}

\definecolor{TableColor}{rgb}{0.882, 0.945, 0.882}

\newcolumntype{C}[1]{>{\centering\arraybackslash}p{#1}}

\def\addlegendimage{\csname pgfplots@addlegendimage\endcsname}

\newcommand{\sectionref}[1]{\hyperref[#1]{Section~\ref{#1}}}

\newcommand{\supplautoref}[1]{\hyperref[#1]{Suppl.~\autoref{#1}}}
\newcommand{\supplsec}[1]{\hyperref[#1]{Suppl. Sec.~\ref{#1}}}

\newcommand{\redcell}[1]{\textcolor{red}{#1}}

\newlength{\lfblockw}
\newlength{\lfgtw}
\newlength{\lfmw}

\usepackage{multirow}
\usepackage{enumitem}
\usepackage[normalem]{ulem}
\usepackage{xspace}
\usepackage{amsmath}

\usepackage{tikz}
\usepackage{pgfplots}
\usepackage{xspace}
\usepackage{amsmath}
\usepackage{colortbl}
\usepackage{enumitem}
\usepackage[normalem]{ulem}
\usepackage{wrapfig}
\usepackage{caption} 
\usepackage{graphicx}
\usepackage{multirow}
\usepackage{etoc}
\usepackage{booktabs}

\usepackage{makecell}
\usepackage{pifont}%

\definecolor{baselinecolor}{rgb}{0.9, 0.9, 1.0}
\usepackage{makecell}

\newcommand{\parag}[1]{\par\noindent\textbf{#1}\hspace{0.5em}}

\title{\ours: End-to-End Learning Predictable Representations for Latent World Models}

\author{Efstathios Karypidis$^{1,2,3}$ \thanks{Part of this work was done during an internship at valeo.ai. Corresponding author: \texttt{e.karypidis@athenarc.gr}. } 
\hspace{0.5em}Spyros~Gidaris$^{2}$ \hspace{0.5em} Nikos~Komodakis$^{1,4,5}$  \vspace{0.5em} 
\\
$^1$Archimedes, Athena Research Center, Greece \hspace{1.0em} $^2$valeo.ai \\
$^3$National Technical University of Athens \hspace{1.0em} $^4$University of Crete \hspace{1.0em} $^5$IACM-Forth
}
\iclrfinalcopy 

\begin{document}

\maketitle

\begin{abstract}
Predicting the future evolution of a scene is a fundamental capability for world modeling. Recent work has shown that operating in the feature space of Vision Foundation Models (VFMs) yields semantically rich representations that support diverse future scene understanding tasks. However, existing approaches rely on two-stage pipelines, where VFM features are first compressed using fixed dimensionality reduction (e.g., PCA) or independently trained autoencoders, and a separate predictor is trained on top of the resulting frozen latent space. This decoupling between representation learning and temporal prediction, as well as approaches that apply predictors directly on raw VFM features, provides no guarantee that the latent space is structured for predictable dynamics. In this work, we propose \ours, an end-to-end framework that jointly learns a latent tokenizer and a flow-based generative dynamics model, explicitly shaping the representation to support temporal predictability. To enable stable joint optimization, we introduce several key design choices that prevent latent collapse and align reconstruction with generative objectives. 
Extensive experiments show that our approach learns more temporally
coherent latent representations and consistently outperforms two-stage
baselines across multiple future scene understanding tasks and prediction
horizons, while eliminating separate training stages, including during
high-resolution adaptation. We provide the implementation code and model weights at \href{https://github.com/Sta8is/Latent-Foresight}{https://github.com/Sta8is/Latent-Foresight}.
\end{abstract}

\section{Introduction}

A central challenge in world modeling is to forecast the future evolution of a scene from its observed past. This capability is critical for autonomous systems, such as robots and vehicles, which must anticipate future events to plan safe and effective actions. A fundamental question underlying this problem is: \textit{what representation should future prediction operate on?}

A common approach is to predict future world states at the pixel level, typically using diffusion models in a two-stage pipeline: an autoencoder (or tokenizer) first compresses RGB frames into a latent space, and a generative model is then trained to forecast future states in this space. While conceptually simple, such latent representations largely preserve low-level visual details, forcing the generative model to capture fine-grained appearance variations that are often irrelevant for downstream decision-making tasks. This results in high computational cost and large model requirements.

Recent work~\citep{karypidis2025dinoforesight, baldassarre2025back, zhou2025dinowm, boduljak2025vfm,sun2026vggt} has shifted toward forecasting in the feature space of Vision Foundation Models (VFMs), such as DINOv2~\citep{oquab2024dinov}, whose representations encode higher-level semantic structure. Operating in this space improves performance on downstream dense prediction tasks (e.g., semantic segmentation, depth estimation, surface normals) while reducing model size and computational cost. Early approaches were discriminative~\citep{karypidis2025dinoforesight, zhou2025dinowm, baldassarre2025back}, relying on deterministic regression to predict future features, often after dimensionality reduction via fixed preprocessing such as PCA~\citep{karypidis2025dinoforesight}. However, deterministic predictors fail to capture the inherent uncertainty of future outcomes, leading to averaged predictions. 

To address this limitation, recent work has introduced generative formulations~\citep{walker2025generalist,boduljak2025vfm, sun2026vggt}, in particular flow-matching models that generate diverse future feature trajectories. Notably, VFMF~\citep{boduljak2025vfm} shows that compressing VFM features with an autoencoder is important for obtaining a well-conditioned latent space suitable for generative modeling. Despite these advances, existing approaches adopt a two-stage training paradigm: a tokenizer is first learned to compress VFM features, and a flow-based forecasting model is subsequently trained on the resulting frozen latent space.

While effective, this decoupled pipeline is suboptimal for generative forecasting. The latent space is optimized for reconstruction, independently of the flow-matching objective, with no guarantee that its geometry is conducive to modeling temporal dynamics. In particular, there is no reason for latent trajectories to be smooth, structured, or predictable under the learned generative flow. This raises a natural question: \textit{can the tokenizer and the flow-matching model be trained jointly, so that the latent representation is shaped directly by the requirements of generative prediction?}

In this work, we answer this question affirmatively. We introduce \ours, a framework that jointly trains a VFM feature tokenizer and a flow-matching model to forecast future latent states. To the best of our knowledge, this is the first approach to enable end-to-end training of both the tokenizer and a generative latent dynamics model in the context of VFM-based world modeling. Transitioning from a two-stage to an end-to-end formulation is, however, non-trivial. Naïvely optimizing both components jointly leads to degenerate solutions in which latent representations collapse to trivial constants, minimizing the generative loss while destroying useful structure.

We show that stable end-to-end training is achievable through simple but critical design choices, including stopping gradients through flow-matching targets and noisy inputs, normalizing the latent space, incorporating a reconstruction loss on denoised latents, and selecting the noise schedule. With these ingredients, the tokenizer and flow-matching model learn complementary objectives, yielding a latent space that is both reconstructive and well-suited for generative forecasting.

Empirically, our approach outperforms prior two-stage methods in forecasting accuracy while preserving reconstruction fidelity, using a single training stage from scratch. Beyond performance gains, our results show that reconstruction and generative objectives can be jointly optimized to shape latent representations for world modeling, yielding a simpler, more principled paradigm.

In summary, our contributions are threefold:
\textbf{(1)} We show
that flow-based latent world models can be trained fully end-to-end, jointly optimizing both the latent encoder and the latent flow-matching dynamics model. This establishes a new paradigm for end-to-end trainable latent world modeling, simplifying training.
\textbf{(2)} To enable stable joint optimization, we introduce several key technical components that address the challenges of coupling latent representation learning with diffusion-based dynamics modeling, preventing collapse and ensuring well-structured latent spaces.
\textbf{(3)} We demonstrate empirically that end-to-end training substantially improves performance over conventional decoupled pipelines, yielding consistent gains across multiple scene understanding tasks and prediction horizons.
\section{Related Work}

\paragraph{Video Prediction-RGB world models} Anticipating future visual states from past observations has long been a central objective in computer vision. Early recurrent approaches based on convolutional LSTMs \citep{nabavi2018future, Xu_2018_CVPR, wang2018eidetic, castrejon2019improved, lee2021video, wu2021motionrnn} established the core temporal modeling paradigm but struggled to produce sharp and temporally consistent predictions. To address this, subsequent work incorporated generative modeling through adversarial training and variational autoencoders \citep{yan2021videogpt, vondrick2016anticipating, babaeizadeh2018stochastic, lee2018stochastic}, improving diversity and visual quality of predicted frames. The introduction of diffusion models \citep{ho2022imagen, ho2022video, harvey2022flexible, gao2022simvp, karypidisre2pix} further advanced prediction fidelity by modeling the full distribution over future frames. Transformer-based architectures extended these gains by capturing long-range temporal dependencies through autoregressive and masked modeling objectives \mbox{\citep{yu2023magvit, yu2024language, gupta2023maskvit, wang2024emu3}}. Most recently, large-scale generative models such as Sora \citep{videoworldsimulators2024} and COSMOS~\citep{ali2025world} have pushed the boundaries of photorealistic video synthesis, though at substantial computational cost. While these methods excel at generating visually realistic futures, they optimize for pixel-level appearance rather than semantic scene understanding, making them poorly suited for the needs of autonomous systems.

\paragraph{Latent Diffusion Models}
Modern video prediction methods commonly adopt latent generative approaches \citep{gupta2023maskvit, yu2023magvit, yu2024language, gao2022simvp}, following a two-stage pipeline in which an autoencoder is first trained to compress frames and a generative model is subsequently trained on the frozen latent space. 
While this reduces computational cost and facilitates optimization, the learned representation is optimized for reconstruction rather than generation. 
Recent works have challenged this paradigm in image generation. 
REPA-E \citep{leng2025repa} enables joint training of a VAE and a diffusion transformer through representation alignment with a frozen vision encoder, accelerating convergence. 
UNITE \citep{duggal2026end} jointly optimizes tokenization and generation through a shared generative encoder, while EOSTok \citep{chu2026} jointly trains an 1D semantic tokenizer and an autoregressive model. 
However, these approaches focus on static image generation. Jointly learning latent representations and generative temporal dynamics for world modeling remains largely unexplored, introducing challenges such as latent collapse and unstable latent geometry. 
Unlike REPA-E, which relies on an additional representation-alignment objective, our work investigates how to achieve stable end-to-end training of a tokenizer and a temporal flow-matching predictor for latent world modeling.

\paragraph{Semantic Future Prediction - Latent World Models} 

Several works have explored predicting future observations directly at the semantic level or in intermediate feature spaces, rather than pixel space~\citep{, luc2017predicting, lin2021predictive, karypidis2025advancing}.
DINO-Foresight \citep{karypidis2025dinoforesight} introduced forecasting in DINOv2 feature space for dense scene understanding, using PCA compression and a masked transformer predictor.
DINO-WM \citep{zhou2025dinowm} leveraged frozen DINOv2 features for action-conditioned planning, while DINO-world \citep{baldassarre2025back} further demonstrated the effectiveness and computational efficiency of DINOv2-based video world models.
V-JEPA \citep{bardes2024revisiting} jointly trains an encoder and predictor through masked spatiotemporal feature prediction, using context from the entire video clip rather than only past observations, with the goal of learning visual representations rather than forecasting future states.
In V-JEPA 2 \citep{assran2025v}, temporal forecasting is instead performed by training a separate predictor on frozen pretrained representations, without jointly optimizing the encoder.
Frozen Forecasting \citep{walker2025generalist} systematically evaluated frozen VFMs for forecasting, highlighting the advantages of video-pretrained representations.
More recently, VFMF \citep{boduljak2025vfm} introduced a separately trained autoencoder and generative flow matching in VFM feature space, while FlowWM~\citep{porcher2026flow} performs flow matching directly on frozen DinoV3 features with a one-step projection for stable high-dimensional training. DeltaTok \citep{kerssies2026frame} proposed a compact tokenizer encoding frame-to-frame feature differences into a single token for efficient generative world modeling. LeWorldModel~\citep{maes2026leworldmodel} also jointly trains an encoder and predictor from scratch, but uses discriminative (MSE-based) prediction of compact global CLS-token representations for planning on synthetic control benchmarks, rather than flow-matching-based generative
prediction of dense spatial features for scene understanding.
Unlike these approaches, our work jointly optimizes a VFM feature tokenizer and a generative temporal predictor end-to-end, explicitly shaping the latent space for temporal predictability.

\section{Methodology}

Our goal is to improve Vision Foundation Model (VFM) feature forecasting~\citep{karypidis2025dinoforesight, zhou2025dinowm, baldassarre2025back, boduljak2025vfm} by learning a latent representation that is explicitly shaped for generative temporal prediction. In contrast to prior approaches that rely on frozen representations~\citep{baldassarre2025back} or separately trained tokenizers~\citep{boduljak2025vfm, karypidis2025dinoforesight}, we propose a unified framework in which the latent space is learned jointly with a flow-based generative model. This allows the representation itself to adapt to the requirements of temporal forecasting rather than being fixed a priori.

\subsection{Model Components}
\label{sec:modules}

\parag{VFM Feature Extraction.}
Following prior work~\citep{karypidis2025dinoforesight}, we extract dense features from multiple intermediate layers of a frozen DINOv2 ViT encoder~\citep{oquab2024dinov},
where early layers capture local structure while deeper layers encode higher-level semantics, leading to improved performance in dense prediction tasks.
Given $L$ selected layers, we extract per-layer feature maps and concatenate them along the channel dimension to obtain a unified representation $\mathbf{F} \in \mathbb{R}^{H \times W \times D_f}$, with $D_f = L \cdot D_{\text{enc}}$,
where $H \times W$ denotes the spatial resolution of the feature grid and $D_{\text{enc}}$ is the per-layer embedding dimension. The objective of VFM forecasting is to predict the evolution of these feature maps over time.

\begin{figure}[t]
\centering
\includegraphics[trim={0cm 0cm 0cm 0cm},width=1\linewidth]{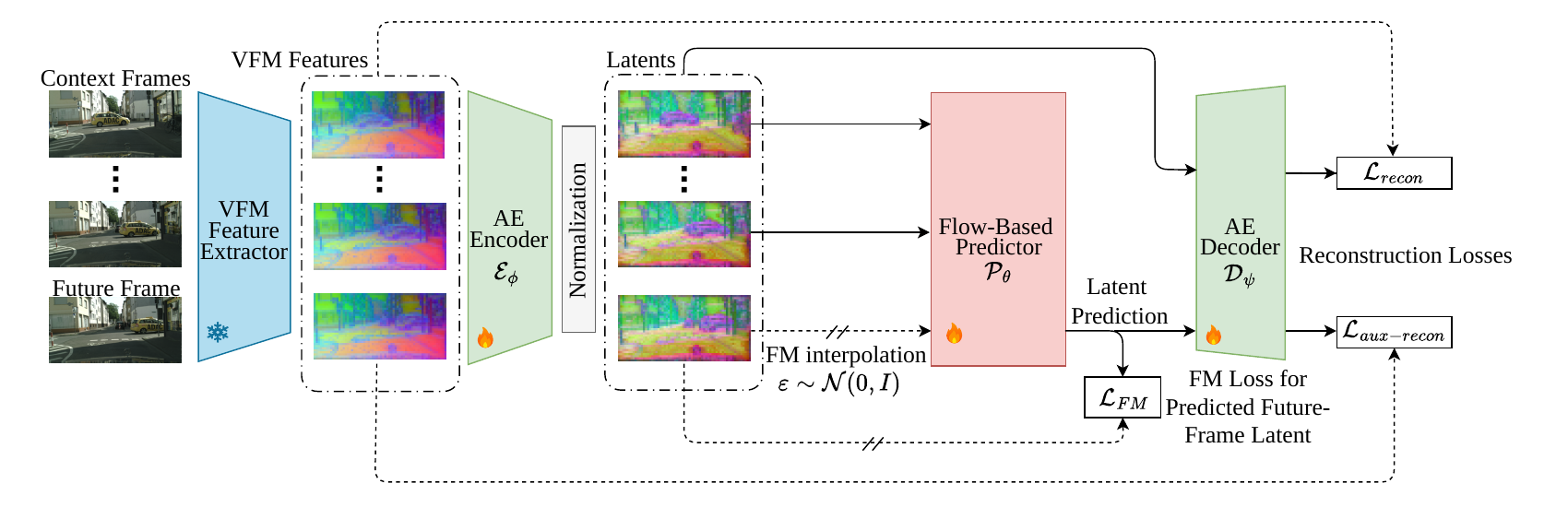}
\vspace{-20pt}
\caption{\textbf{Overview of \ours.} We jointly train end-to-end a VFM tokenizer encoder $\mathcal{E}_\phi$, decoder $\mathcal{D}_\psi$, and flow-based predictor $\mathcal{P}_\theta$, while keeping the VFM backbone frozen. The encoder compresses per-frame features into a compact predictable latent space, followed by a normalization layer. The decoder is supervised by $\mathcal{L}_{\text{recon}}$. The predictor takes context latents and a noisy interpolation of the future latent as input, trained with $\mathcal{L}_{\text{FM}}$. Stop-gradient operations on both the future latent and the noisy input prevent collapse. $\mathcal{L}_{\text{aux-recon}}$ supervises the predicted features against the ground truth.}
\label{fig:overview}
\vspace{-10pt}
\end{figure}

\parag{VFM Feature Tokenizer.}
The resulting feature maps are high-dimensional, often exceeding several thousand channels per spatial location, which makes direct generative modeling challenging. 
To address this, prior work has introduced dimensionality reduction schemes such as PCA or learned autoencoders.
We follow the autoencoder-based formulation and introduce an encoder $\mathcal{E}_\phi$ and decoder $\mathcal{D}_\psi$, parameterized by $\phi$ and $\psi$ (implemented as light-weight vision transformers, with two transformer blocks each). The encoder maps each frame-level feature map into a compact latent representation 
$\mathbf{z} = \mathcal{E}_\phi(\mathbf{F}) \in \mathbb{R}^{H \times W \times C}$,
with $C \ll D_f$, while the decoder reconstructs the original VFM features as
$\hat{\mathbf{F}} = \mathcal{D}_\psi(\mathbf{z}) \in \mathbb{R}^{H \times W \times D_f}$.
The autoencoder is trained with a per-frame reconstruction objective combining Euclidean and cosine similarity terms, following \citep{SemVAE}:
\begin{equation}
    \mathcal{L}_{\text{recon}} = \frac{1}{N} \sum_{i=1}^{N} \left\| \mathcal{D}_\psi\!\left(\mathcal{E}_\phi(\mathbf{F}_i)\right) - \mathbf{F}_i \right\|_2^2 + \left(1 - \cos\!\left(\mathcal{D}_\psi\!\left(\mathcal{E}_\phi(\mathbf{F}_i)\right),\, \mathbf{F}_i\right)\right).
\end{equation}

\parag{Flow-Based Latent Forecasting.}
We model temporal dynamics in the latent space using a flow-matching formulation. Given a video sequence of $N = N_c + N_p$ frames, we focus on next-frame prediction ($N_p = 1$), where the goal is to forecast the future latent $\mathbf{z}_f = \mathbf{z}_{N_c+1}$ conditioned on the context latents $\mathbf{z}_c = \{\mathbf{z}_1, \dots, \mathbf{z}_{N_c}\}$. Longer prediction horizons are obtained autoregressively.

We define a continuous interpolation between the target latent and Gaussian noise $\boldsymbol{\varepsilon} \sim \mathcal{N}(0, I)$ as
\begin{equation}
    \mathbf{z}_f^{\,t} = t\, \mathbf{z}_f + (1 - t)\, \boldsymbol{\varepsilon}, \quad t \in [0,1],
\end{equation}
where $t = 1$ corresponds to clean data and $t = 0$ corresponds to pure noise. 

We adopt the $\mathbf{x}$-prediction parameterization of JiT~\citep{li2025back}, where the predictor $\mathcal{P}_\theta$ directly estimates the clean latent:
$\hat{\mathbf{z}}_f = \mathcal{P}_\theta(\mathbf{z}_f^{\,t}, t, \mathbf{z}_c)$.
This parameterization has demonstrated improved stability and performance in high-dimensional generative modeling.
The corresponding flow-matching objective is formulated in velocity form:
\begin{equation}
    \mathcal{L}_{\text{FM}} =
    \mathbb{E}_{t \sim p(t),\, \boldsymbol{\varepsilon} \sim \mathcal{N}(\mathbf{0}, \mathbf{I})}
    \left[
        \left\| \hat{\mathbf{v}} - \mathbf{v} \right\|^2
    \right],
\end{equation}
where $\mathbf{v} = \mathbf{z}_f - \boldsymbol{\varepsilon} = (\mathbf{z}_f - \mathbf{z}_f^{\,t}) / (1 - t)$
and $\hat{\mathbf{v}} = (\hat{\mathbf{z}}_f - \mathbf{z}_f^{\,t}) / (1 - t)$
denote the target and predicted velocities, respectively. The sampling distribution $p(t)$ follows a logit-normal distribution with $\text{logit}(t) \sim \mathcal{N}(\mu, \sigma^2)$, controlling the concentration of training noise levels.
\subsection{End-to-End Learning}
\label{sec:e2e}

Existing latent world models operate on frozen representations, either directly from a VFM encoder or from a separately trained tokenizer. In contrast, our framework jointly learns the tokenizer and the flow-based predictor, allowing the latent space to be shaped by the requirements of generative forecasting. This coupling ensures that the encoder does not merely compress features, but produces representations that support predictable temporal evolution under the flow model.

However, joint optimization introduces a non-trivial training dynamic. Without appropriate constraints, the model may converge to degenerate solutions where latent representations collapse to trivial constants that minimize prediction error. 
The goal is therefore to prevent collapse while encouraging latents that are both reconstructive and temporally structured.

\parag{Latent Normalization.}
Since the flow model interpolates between latent features and Gaussian noise, their scales must be compatible. We normalize the encoder $\mathcal{E}_\phi$ outputs using Batch Normalization~\citep{ioffe2015batch} without affine parameters, enforcing zero-mean unit-variance latents. This prevents either signal or noise dominance during interpolation.
We additionally experimented with Layer Normalization~\citep{ba2016layer} and regularization objectives based on KL divergence~\citep{kingma2013auto} and SIGReg~\citep{SigReg}.

\parag{Stop-Gradient Strategy.}
To avoid trivial solutions while maintaining meaningful gradient flow, we apply stop-gradient operations in the flow objective. The velocity target is computed as
\begin{equation}
    \mathbf{v} = \text{sg}(\mathbf{z}_f) - \boldsymbol{\varepsilon},
\end{equation}
preventing gradients from affecting the target branch. Additionally, the noisy input is detached,
\begin{equation}
    \hat{\mathbf{z}}_f = \mathcal{P}_\theta(\text{sg}(\mathbf{z}_f^{\,t}), t, \mathbf{z}_c),
\end{equation}
ensuring that gradients flow only through the context latents. 
As a result, the prediction objective shapes the encoder primarily through the context representations, encouraging latent features that support predictable temporal dynamics while preventing shortcut solutions in the noisy input latents.

\parag{Prediction Reconstruction Regularization.}
To preserve information in the latent space, we maintain the reconstruction objective from the tokenizer. In addition, we introduce an auxiliary reconstruction loss applied to predicted latents:
\begin{equation}
\mathcal{L}_{\text{aux-recon}} =
\left\|
\mathcal{D}_\psi(\hat{\mathbf{z}}_{f}) - \mathbf{F}_{N_c+1}
\right\|_2^2
+ 1 - \cos(\mathcal{D}_\psi(\hat{\mathbf{z}}_{f}), \mathbf{F}_{N_c+1}).
\end{equation}
This term helps reduce the train-test mismatch between encoded and generated latents, while remaining secondary to the main objectives.

\parag{Full Objective.}
The overall training objective is
\begin{equation}
    \mathcal{L} = \mathcal{L}_{\text{recon}} + \mathcal{L}_{\text{aux-recon}} + \mathcal{L}_{\text{FM}},
\end{equation}
where all components are optimized jointly over the encoder $\mathcal{E}_\phi$, decoder $\mathcal{D}_\psi$, and predictor $\mathcal{P}_\theta$.

\section{Experiments}
\label{sec:experiments}

\subsection{Experimental Setup}
\label{sec:setup}          

\parag{Data.}
We evaluate our approach on two urban driving datasets, Cityscapes~\citep{Cordts_2016_CVPR} and nuScenes~\citep{nuscenes}, as well as Kubric~\citep{Kubric}, a synthetic multi-object dynamics benchmark, following the evaluation protocols of~\citep{karypidis2025dinoforesight, boduljak2025vfm}. To assess scalability, we additionally train a variant, \oursplus, on the combined Cityscapes, nuScenes, and CoVLA~\citep{covla} data. More dataset details in Appendix~\ref{app:datasets}.

\parag{Implementation Details.}
By default, we use DINOv2-Reg with ViT-B/14 as the visual encoder, with features extracted from $L=4$ intermediate layers and concatenated to form a $D_f = 3072$-dimensional representation per token. Our learned autoencoder $\mathcal{E}_\phi, \mathcal{D}_\psi$ incorporates spatial Rotary Position Embeddings (RoPE) and projects each per-frame feature map $\mathbf{F}_i \in \mathbb{R}^{H \times W \times D_f}$ to a compact latent $\mathbf{z}_i \in \mathbb{R}^{H \times W \times C}$ with bottleneck dimension $C = 256$. The predictor builds upon the DINO-Foresight architecture~\citep{karypidis2025dinoforesight}, extended with RMSNorm~\citep{rmsnorm}, QK-normalization~\citep{henry2020query}, and SwiGLU~\citep{shazeer2020glu} activations for improved training stability at scale, along with spatiotemporal RoPE, and is conditioned on the flow-matching noise level $t$ via adaLN-zero~\citep{peebles2023scalable}. For evaluation, we train frozen DPT heads~\citep{ranftl2021vision} for semantic segmentation, depth prediction, and surface normal estimation. In addition to our default setting, we report \textbf{Latent-Foresight+}, a scaled variant trained on additional data with a longer training schedule, to demonstrate the scalability of our approach. Full architecture, optimization, and training details are provided in \hyperref[sec:appendix-implemdetails]{Appendix Subsection~\ref*{sec:appendix-implemdetails}}.

\parag{Evaluation Metrics.}
We evaluate future prediction quality across multiple scene understanding tasks. For \textbf{semantic segmentation}, we report mean Intersection over Union (mIoU) computed over all classes (ALL) and restricted to movable object classes (MO). For \textbf{depth prediction}, we report mean Absolute Relative Error (AbsRel$\downarrow$) and threshold accuracy ($\delta_1\uparrow$). For \textbf{surface normals}, we report mean angular error (m$\downarrow$) and the percentage of pixels with angular error below $11.25^\circ$ ($11.25^\circ\uparrow$). In ablation studies, we additionally report the cosine similarity between reconstructed or predicted features and the ground-truth VFM features. On Cityscapes, we evaluate \textbf{short-term} and \textbf{mid-term} prediction; on nuScenes, which captures more static scenes with slower dynamics, we additionally evaluate a \textbf{longer-term} horizon. On Kubric, we follow a separate step-based protocol for comparison with prior work. Full metric definitions, movable-object classes, and per-dataset horizon lengths are provided in \hyperref[sec:appendix-evalmetrics]{Appendix Subsection~\ref*{sec:appendix-evalmetrics}.}

\parag{Baselines.}
The \textbf{\texttt{Oracle}} baseline directly accesses the ground-truth future frame, providing an upper bound on performance. 
We additionally compare against \textbf{\texttt{VISTA}}~\citep{gao2024vista}, a state-of-the-art latent video diffusion world model comprising 2.5 billion parameters and trained on 1,740 hours of driving video. VISTA generates future RGB frames conditioned on past observations without action inputs. The generated frames are subsequently processed by the DINOv2-Reg encoder and frozen DPT heads for evaluation.
Finally, we train and evaluate two-stage baselines in which the autoencoder-based tokenizer is first optimized independently and subsequently frozen while training the flow-based predictor. This setting follows a pipeline conceptually similar to VFMF~\citep{boduljak2025vfm} and enables a direct comparison with our end-to-end formulation.

\subsection{Comparative Results}

\newcommand{\graycell}[1]{\textcolor{gray}{#1}}
\begin{table}[t]
\scriptsize
\centering
\setlength{\tabcolsep}{3pt}
\caption{\textbf{Comparison with state-of-the-art on high-resolution VFM forecasting ($448\times 896$).} Methods that do not support a task are marked with `-'. 
ALL: mIoU over all classes. MO: mIoU over movable-object classes. 
We compare \ours with \texttt{DINO-Foresight}~\citep{karypidis2025dinoforesight} and \texttt{DeltaTok}~\citep{kerssies2026frame} (for which results are taken from the original paper). We also include the pixel-level future generation method \texttt{VISTA}~\citep{gao2024vista}. \texttt{VISTA}$_{ft}$ denotes the \texttt{VISTA} model fine-tuned on Cityscapes. \texttt{\oursplus} denotes our \ours model trained on additional data for a longer training schedule.}
\vspace{-10pt}
\begin{tabular}{l
    C{0.7cm} C{0.7cm} C{0.7cm} C{0.7cm}
    C{0.8cm} C{0.8cm} C{1.0cm} C{1.0cm}
}
\toprule
\multirow{3}{*}[-1.0ex]{\Th{Method}}
& \multicolumn{4}{c}{\Th{Semantic Segmentation}}
& \multicolumn{2}{c}{\Th{Depth}}
& \multicolumn{2}{c}{\Th{Surface Normals}} \\
\cmidrule(lr){2-5} \cmidrule(lr){6-7} \cmidrule(lr){8-9}
& \multicolumn{2}{c}{\Th{Short}} & \multicolumn{2}{c}{\Th{Mid}}
& \Th{Short} & \Th{Mid}
& \Th{Short} & \Th{Mid} \\
\cmidrule(lr){2-3} \cmidrule(lr){4-5} \cmidrule(lr){6-6} \cmidrule(lr){7-7} \cmidrule(lr){8-8} \cmidrule(lr){9-9}
& ALL & MO & ALL & MO
& $\delta_{1}$ & $\delta_{1}$ & 11.25$^\circ$ & 11.25$^\circ$ \\

\midrule

\graycell{\texttt{Oracle}} & \graycell{77.1} & \graycell{77.3} & \graycell{77.1} & \graycell{77.3} & \graycell{89.6} & \graycell{89.6} & \graycell{96.3} & \graycell{96.3} \\
\texttt{VISTA}$_{ft}$ & 64.9 & 62.1 & 53.9 & 51.0 & 86.4 & 82.8 & 93.0 & 90.0 \\
\texttt{Dino-Foresight} & 71.8 & 71.7 & 59.8 & 57.6 & 88.6 & 85.4 & 94.4 & 91.3 \\
\texttt{DeltaTok} & 72.1 & - & 60.0 & - & 88.5 & 85.6 & - & - \\
\rowcolor{baselinecolor} \ours  & \underline{72.8} & \underline{72.6} & \underline{61.7} & \underline{59.9} & \underline{88.9} & \underline{86.6} & \underline{95.0} & \underline{92.0} \\
\rowcolor{baselinecolor} \oursplus  &  \textbf{73.3} &  \textbf{72.9} &  \textbf{63.3} & \textbf{61.9} & \textbf{88.9} & \textbf{86.9} & \textbf{95.2} & \textbf{92.4}  \\
\bottomrule
\end{tabular}
\label{tab:comparison_with_prior_work}
\vspace{-10pt}
\end{table}

\parag{Comparison with State-of-the-Art on Cityscapes.}
Table~\ref{tab:comparison_with_prior_work} compares \ours against state-of-the-art VFM forecasting methods (all metrics are reported in \hyperref[tab:ablation_sota_full]{Appendix Table~\ref*{tab:ablation_sota_full}}).
Our generative framework with jointly learned latents consistently outperforms the discriminative \texttt{DINO-Foresight}~\citep{karypidis2025dinoforesight} across all metrics and prediction horizons.
The gains increase at longer horizons, highlighting the benefits of our approach for long-term autoregressive forecasting.
Our method also surpasses the recent \texttt{DeltaTok}~\citep{kerssies2026frame}.
The pixel-level future generation method \texttt{VISTA}~\citep{gao2024vista}, despite its significantly larger scale, underperforms across all tasks and horizons, highlighting the effectiveness of semantic feature forecasting for future scene understanding.
Finally, scaling our approach (\oursplus) with more data and longer training consistently improves all metrics, demonstrating its scalability.
We further compare against the generative \texttt{VFMF}~\citep{boduljak2025vfm}, a two-stage flow-matching approach, at $224\times448$ resolution in Table~\ref{tab:ablation_ae} (its only available checkpoint).
We also include an ablated two-stage variant of our method (Two-Stage with AE), which isolates the effect of end-to-end training.
Our method consistently outperforms \texttt{VFMF}, with additional comparisons on Kubric in \autoref{tab:kubric_1column}.

\begin{table}[t]
\caption{\textbf{Impact of end-to-end latent learning.}  
Comparison between our end-to-end formulation, two-stage pipelines based on PCA or learned autoencoders, and forecasting directly on raw VFM features. 
We also include the two-stage flow-matching method \texttt{VFMF}~\citep{boduljak2025vfm}.
We report reconstruction fidelity and forecasting performance at short- and mid-term horizons. \textbf{Cos Sim} denotes cosine similarity between predicted/reconstructed features and ground-truth VFM features. Results are reported at low and high resolution.
For \texttt{VFMF}$^\dagger$, we evaluate the released checkpoint by sampling 32 flow-matching trajectories with 10 ODE steps each and averaging predicted latents. 
End-to-end latent learning consistently improves results while maintaining stable optimization.
}
\vspace{-10pt}
\scriptsize
\centering
\setlength{\tabcolsep}{2pt}
\resizebox{\textwidth}{!}{%
\begin{tabular}{l @{\hspace{1.5em}} C{0.5cm} @{\hspace{1.5em}} C{1.4cm} C{0.75cm} C{0.75cm} @{\hspace{1.5em}} C{1.4cm} C{0.75cm} C{0.75cm} @{\hspace{1.5em}} C{1.4cm} C{0.75cm} C{0.75cm}}
\toprule
\multirow{2}{*}[-0.5ex]{\Th{Method}}
& \multirow{2}{*}[-0.5ex]{\Th{Dim}}
& \multicolumn{3}{c}{\Th{Reconstruction}}
& \multicolumn{3}{c}{\Th{Prediction (Short-Term)}}
& \multicolumn{3}{c}{\Th{Prediction (Mid-Term)}} \\
\cmidrule(lr){3-5} \cmidrule(lr){6-8} \cmidrule(lr){9-11}
& & Cos Sim$\uparrow$ & ALL$\uparrow$ & MO$\uparrow$
& Cos Sim$\uparrow$ & ALL$\uparrow$ & MO$\uparrow$
& Cos Sim$\uparrow$ & ALL$\uparrow$ & MO$\uparrow$ \\
\midrule
\multicolumn{11}{c}{\textit{Low Resolution $224 \times 448$}} \\
Raw VFM Features          & 3072 & 1.0 & 68.12 & 66.81 & 0.968 & 64.02 & 62.78 & 0.933 & 54.22 & 50.77 \\

Two Stage with PCA  & 1152 & 0.977 & 65.63 & 61.97 & 0.955 & 61.56 & 57.93 & 0.921 & 52.49 & 47.60 \\
Two Stage with PCA  & 256 & 0.952 &  51.19 & 37.18 & 0.940 & 47.85 & 33.63 & 0.914 & 43.10 & 29.46 \\

Two-Stage with AE & 256 & \textbf{0.989} & 67.96 & 66.88 & 0.967 & 64.69 & 62.57 & 0.938 & 56.31 & 53.03 \\

\texttt{VFMF}$^\dagger$ & 16 & 0.975 & 66.11 & 64.96 & 0.959 & 62.01 & 60.00 & 0.931 &  51.54 & 45.63  \\
\rowcolor{baselinecolor} End-to-End (Ours) & 256 & \textbf{0.989} &  \textbf{68.29} & \textbf{67.26} & \textbf{0.972} & \textbf{65.45} & \textbf{64.10} & \textbf{0.944} & \textbf{56.84} & \textbf{53.86} \\
\midrule
\multicolumn{11}{c}{\textit{High Resolution $448 \times 896$}} \\
Two Stage with AE & 256 & \textbf{0.991} & \textbf{77.05} & 77.30 & 0.967  & 71.43 & 70.36 & 0.930 & 60.18 & 58.05 \\
\rowcolor{baselinecolor} End-to-End (Ours) & 256 & \textbf{0.991} & 77.04 & \textbf{77.40} & \textbf{0.972} &\textbf{72.77} & \textbf{72.57} & \textbf{0.937} & \textbf{61.74} & \textbf{59.92}   \\
\bottomrule
\end{tabular}}
\label{tab:ablation_ae}
\vspace{-7pt}
\end{table}

\begin{table}[!t]
\footnotesize
\centering
\caption{\textbf{Comparison across tasks at nuScenes with resolution $448\times 896$}. 
We
show cosine similarity between predicted and ground-truth features, and depth estimation performance ($\delta_{1}$ accuracy, AbsRel error), across mid-term (9 frames, 0.75s), long-term (18 frames, 1.5s), and longer-term (27 frames, 2.25s) horizons. 
The last two rows evaluate zero-shot generalization: models trained on Cityscapes and directly evaluated on nuScenes.} 
\vspace{-10pt}
\setlength{\tabcolsep}{2pt}
\resizebox{0.8\textwidth}{!}{%
\begin{tabular}{lccccccccc}
\toprule
 \multirow{3}{*}[-1.0ex]{\Th{Method}} & \multicolumn{3}{c}{\Th{Cos Sim}$\uparrow$} & \multicolumn{3}{c}{\Th{$\delta_1$}$\uparrow$} & \multicolumn{3}{c}{\Th{AbsRel}$\downarrow$} \\
\cmidrule(lr){2-4} \cmidrule(lr){5-7} \cmidrule(lr){8-10}

& \Th{Mid} & \Th{Long} & \Th{Longer} & \Th{Mid} & \Th{Long} & \Th{Longer} & \Th{Mid} & \Th{Long} & \Th{Longer} \\
\midrule
\graycell{\texttt{Oracle}} & \graycell{-} & \graycell{-} & \graycell{-} & \graycell{84.0} & \graycell{84.0} & \graycell{84.0} & \graycell{.138} & \graycell{.138} & \graycell{.138} \\
\midrule
{Dino-Foresight} & 0.924 & 0.894 & 0.868 & 76.8 & 73.1 & 69.5 & 0.321 & 0.377 & 0.368 \\
{Two-Stage with AE} & 0.927 & 0.888 & 0.857 & 75.9 & 72.0 & 68.2 & 0.242 & 0.328 & 0.390 \\
{\ours} & \textbf{0.936} & \textbf{0.903} & \textbf{0.874} & \textbf{80.8} & \textbf{76.4} & \textbf{72.0} & \textbf{0.206} & \textbf{0.267} & \textbf{0.310} \\
\midrule
{Dino-Foresight (zero-shot)} & 0.905 & 0.865 & 0.837 & 74.8 & 69.5 & 66.0 & 0.349 & 0.424 & 0.415 \\
{\ours (zero-shot)} & 0.920 & 0.876 & 0.848 & 77.2 & 71.8 & 68.2 & 0.306 & 0.387 & 0.392 \\
\bottomrule
\end{tabular}}
\label{tab:nuscenes}
\vspace{-10pt}
\end{table}

\parag{Impact of End-to-End Latent Learning.}
Table~\ref{tab:ablation_ae} evaluates the impact of end-to-end latent learning on reconstruction fidelity and forecasting performance, comparing our approach against raw VFM features and two-stage pipelines based on PCA or learned autoencoders (AE).
Among these baselines, the learned autoencoder (Two-Stage with AE) achieves the strongest overall performance, outperforming both PCA-based compression and direct prediction on raw VFM features.
In contrast, PCA at the same latent dimensionality ($C=256$) fails to preserve sufficient feature information, substantially degrading both reconstruction and forecasting performance.
These results highlight the importance of learned compression for generative forecasting.
However, independently training the autoencoder leaves the latent space fixed during predictor training, limiting its adaptation to the forecasting objective.
Our end-to-end formulation addresses this limitation by jointly optimizing the tokenizer and flow-based predictor, consistently improving forecasting performance while preserving reconstruction fidelity and maintaining stable training.
The gains are particularly pronounced for movable-object classes (MO), suggesting that jointly learned representations better capture dynamic scene content.
Moreover, the improvements over two-stage training become more pronounced after high-resolution fine-tuning ($448\times896$), as shown on Cityscapes in \autoref{tab:ablation_ae} and nuScenes in \autoref{tab:nuscenes}.
Beyond these performance gains, our single-stage end-to-end approach simplifies high-resolution adaptation, avoiding the separate fine-tuning of the autoencoder and subsequent fine-tuning of the predictor required by two-stage pipelines.

\parag{Generalization Across Driving and Non-Driving Datasets and Longer Prediction Horizons.}
To verify that the benefits of end-to-end latent learning extend beyond Cityscapes, we evaluate on nuScenes, a second real-world driving dataset, and Kubric, a synthetic non-driving benchmark of multi-object dynamics.
Table~\ref{tab:nuscenes} reports nuScenes results across mid-term (9 frames, 0.75s), long-term (18 frames, 1.5s), and longer-term (27 frames, 2.25s) horizons, including zero-shot transfer from models trained only on Cityscapes.
\ours consistently outperforms both DINO-Foresight and our Two-Stage with AE baseline across all metrics and horizons, with gains persisting even at 27 frames, demonstrating its effectiveness for long autoregressive rollouts.
In the zero-shot setting, \ours outperforms DINO-Foresight across all metrics, suggesting that the learned latent representations transfer across driving datasets.
Table~\ref{tab:kubric_1column} evaluates generalization beyond driving on Kubric.
With a single generation, \ours achieves higher foreground IoU than VFMF at both 1-step and 8-step horizons, even compared to its 32-generation setting, while requiring substantially less inference compute.
These results demonstrate that the benefits of end-to-end latent learning extend beyond autonomous driving.

\subsection{Experimental analysis}
\label{sec:analysis}

\begin{wraptable}[13]{r}{0.6\linewidth}
\vspace{-13pt}
\caption{\textbf{Comparison with VFMF on Kubric} (224$\times$224), 
following the VFMF evaluation protocol (4 conditioning frames, timestep of 2). VFMF results are averaged over 32 generated samples following the original paper; \ours requires only a single generation.  VFMF* denotes results as originally reported in the paper.}
\vspace{-5pt}

\centering
\setlength{\tabcolsep}{4pt}
\resizebox{\linewidth}{!}{%
\begin{tabular}{lcccccc}
\toprule
& \multicolumn{3}{c}{\Th{1-step}} & \multicolumn{3}{c}{\Th{8-step}} \\
\cmidrule(lr){2-4} \cmidrule(lr){5-7}
\Th{Method} & \Th{BG}$\uparrow$ & \Th{ALL}$\uparrow$ & \Th{MO}$\uparrow$ & \Th{BG}$\uparrow$ & \Th{ALL}$\uparrow$ & \Th{MO}$\uparrow$ \\
\midrule
VFMF (1 gen)  & 97.71 & 88.19 & 78.66 & 90.34 & 59.53 & 28.71 \\
VFMF*(32 gen)         & --    & --    & --    & 91.62   & 61.74 & 31.86 \\
VFMF (32 gen) & 98.09 & 89.99 & 81.88 & \textbf{91.95} & 62.07 & 32.19 \\
\ours (1 gen) & \textbf{98.32} & \textbf{91.16} & \textbf{83.99} & 91.64 & \textbf{63.47} & \textbf{35.29} \\
\bottomrule
\end{tabular}}
\label{tab:kubric_1column}
\end{wraptable}

\paragraph{Training Recipe Ablation.}
\autoref{tab:ablation_training} evaluates the key components of our end-to-end training strategy.
Our full model (a) achieves the strongest overall performance across reconstruction and forecasting.
Removing RoPE from the autoencoder (b) has only a minor impact, whereas removing the auxiliary reconstruction loss $\mathcal{L}_{\text{aux-recon}}$ (c) consistently degrades forecasting performance, particularly at longer horizons.
This supports the role of auxiliary reconstruction in reducing the train--test mismatch between encoded and predicted latents.
\emph{Latent normalization is essential for stable training.}
Replacing Batch Normalization with Layer Normalization (d) preserves competitive reconstruction quality but degrades forecasting performance, particularly at longer horizons.
KL regularization toward a unit Gaussian (e) substantially degrades both reconstruction and forecasting, whereas SIGReg (f) performs comparably to Batch Normalization.
We therefore retain Batch Normalization for its simplicity and effectiveness.
Removing latent normalization entirely (g) leads to training collapse, highlighting the importance of controlling the latent distribution during end-to-end flow matching.
\emph{Finally, the stop-gradient operations play distinct roles.}
Removing the stop-gradient on the target latent (h) causes training collapse, as it prevents the prediction objective from destabilizing the encoder.
Removing the stop-gradient on the noisy latent (i) preserves training stability but degrades forecasting performance, suggesting that gradients propagated through the noisy interpolation interfere with learning predictive representations.

\begin{table}[b]
\vspace{-7pt}
\caption{\textbf{Ablation of the end-to-end training strategy.}  
We analyze the contribution of the main components enabling stable end-to-end latent learning, including the auxiliary reconstruction objective, latent normalization scheme, and stop-gradient operations used in the flow-matching objective. All results are reported at resolution $224 \times 448$. Removing latent normalization or the stop-gradient on the target latent leads to training collapse, highlighting their importance for stable optimization.  
}
\vspace{-10pt}
\scriptsize
\centering
\setlength{\tabcolsep}{1pt}
\resizebox{\textwidth}{!}{%
\begin{tabular}{cl @{\hspace{1.5em}} C{1.4cm} C{0.75cm} C{0.75cm} @{\hspace{1.5em}} C{1.4cm} C{0.75cm} C{0.75cm} @{\hspace{1.5em}} C{1.4cm} C{0.75cm} C{0.75cm}}
\toprule
& \multirow{2}{*}[-0.5ex]{\Th{Method}}
& \multicolumn{3}{c}{\Th{Reconstruction}}
& \multicolumn{3}{c}{\Th{Prediction (Short-Term)}}
& \multicolumn{3}{c}{\Th{Prediction (Mid-Term)}} \\
\cmidrule(lr){3-5} \cmidrule(lr){6-8} \cmidrule(lr){9-11}
& & Cos Sim$\uparrow$ & ALL$\uparrow$ & MO$\uparrow$
& Cos Sim$\uparrow$ & ALL$\uparrow$ & MO$\uparrow$
& Cos Sim$\uparrow$ & ALL$\uparrow$ & MO$\uparrow$ \\

\midrule
\rowcolor{baselinecolor} (a) & \ours (Ours) & \textbf{0.989} &  \textbf{68.29} & \textbf{67.26} & \textbf{0.972} & \textbf{65.45} & 64.10 & \textbf{0.944} & \textbf{56.84} & \textbf{53.86} \\
(b) & \,\, w/o RoPE in AE & \textbf{0.989} & 68.22 & \textbf{67.26} & 0.971 & 65.42 & \textbf{64.37} & \textbf{0.944} & 56.60 & 53.42\\

(c) & \,\, \,\, \,\, w/o Auxiliary Reconstruction ($\mathcal{L}_{\text{aux-recon}}$) & 0.988 & 68.27 & 67.19 & 0.967 & 65.09 & 63.43 & 0.937 & 56.55 & 53.36 \\
(d) & \,\, \,\, \,\, BatchNorm $\rightarrow$ LayerNorm & 0.989 & 68.27 & 67.22 & 0.970 & 64.70 & 63.53 & 0.942 & 55.88 & 52.66 \\
(e) & \,\, \,\, \,\, BatchNorm $\rightarrow$ KL Regularization & 0.968 & 64.43 & 64.50 & 0.955 & 61.33 & 61.31 & 0.913 & 46.64 & 41.78 \\
(f) & \,\ \,\, \,\,\,  BatchNorm $\rightarrow$ SIGReg & 0.988 & 68.19 & 67.03 & 0.971 & \textbf{65.45} & 64.22 & \textbf{0.944} & 56.60 & 53.46 \\
(g) & \,\, \,\, \,\, \redcell{w/o Latent Normalization} & \multicolumn{9}{c}{\textit{\textcolor{red}{Training Collapse}}} \\
(h) & \,\, \,\, \,\, \redcell{w/o Stop-Grad on Target Latent ($\text{sg}(\mathbf{z}_{f})) - \boldsymbol{\varepsilon}$} & \multicolumn{9}{c}{\textit{\textcolor{red}{Training Collapse}}} \\
(i) &  \,\, \,\, \,\, w/o Stop-Grad on Noisy Latent ($\text{sg}(\mathbf{z}_f^{\,t})$) & 0.987 & 67.97 & 66.52 & 0.966 & 63.17 & 61.47 & 0.937 & 54.48 & 51.33 \\
\bottomrule
\end{tabular}}
\label{tab:ablation_training}
\vspace{-10pt}
\end{table}

\begin{table*}[t]
\begin{minipage}[t]{0.45\linewidth}
    \caption{\textbf{Ablation of Bottleneck Dimensionality.} We vary the latent dimension $C$ of the autoencoder and report reconstruction fidelity and prediction quality at short- and mid-term horizons. All results are reported at $224 \times 448$ resolution and do not include RoPE in the AE.}
    \vspace{-10pt}
    \resizebox{\linewidth}{!}{%
    \scriptsize
\centering
\setlength{\tabcolsep}{4pt}
\begin{tabular}{lccc}
\toprule
& \Th{Reconstruction} & \multicolumn{2}{c}{\Th{Prediction MO}$\uparrow$}\\
\Th{$C$} & \Th{Cos Sim}$\uparrow$ & \Th{Short-term} & \Th{Mid-term} \\
\midrule
32   & 0.980 & 59.93 & 49.82 \\
64   & 0.984 & 62.48 & 54.03 \\
128  & 0.986 & 64.03 & 54.12 \\
\rowcolor{baselinecolor} 256  & 0.989 & 64.37 & 53.42 \\
512  & 0.990 & 64.00 & 54.24 \\
1152 & 0.991 & 63.98 & 51.99 \\
\bottomrule
\end{tabular}
\label{tab:ablation_dimensionality_compact}
    }
\end{minipage}
\hfill
\begin{minipage}[t]{0.52\linewidth}
    \caption{\textbf{Ablation of Noise Level Distribution $p(t)$.} We compare uniform sampling and logit-normal distributions with different mean and standard deviation parameters for training noise levels. All results are reported at $224 \times 448$ resolution, do not include RoPE in the autoencoder, and use a latent dimension of $C=512$.}
    \vspace{-10pt}
    \resizebox{\linewidth}{!}{%
    \scriptsize
\centering
\setlength{\tabcolsep}{4pt}
\begin{tabular}{lccc}
\toprule
& \Th{Recon.} & \multicolumn{2}{c}{\Th{Prediction MO}$\uparrow$}\\
\Th{Noise Distribution}  & \Th{Cos Sim}$\uparrow$ & \Th{Short-term} & \Th{Mid-term} \\
\midrule
Uniform                 & 0.990 & 63.11 & 51.00 \\
Logit-Normal (-0.8,0.8) & 0.990 & 62.75 & 50.89 \\
Logit-Normal (-2,0.8)   & 0.990 & 64.40 & 53.42 \\
\rowcolor{baselinecolor} Logit-Normal (-2,1.5) & 0.990 & 64.00 & 54.24 \\
\bottomrule
\end{tabular}
\label{tab:ablation_noisesched_compact}
    }
\end{minipage}
\end{table*}

\begin{figure}[ht]
    \centering
    \resizebox{\linewidth}{!}{%
{
\small
\centering
\newcommand{\resultsfignew}[1]{\includegraphics[width=0.145\textwidth,valign=c]{#1}}
\setlength{\tabcolsep}{1pt}
\renewcommand{\arraystretch}{0.9}

\newcommand{\rlscene}{figs/scene_95_long}
\newcommand{\rgbctxfig}[1]{\resultsfignew{\rlscene/rgb_#1}}
\newcommand{\segmctxfig}[1]{\resultsfignew{\rlscene/lf_scale/#1}}
\newcommand{\predfig}[2]{\resultsfignew{\rlscene/#1/t_plus_#2}}
\newcommand{\rowname}[1]{\parbox[c]{1.3cm}{\centering\scriptsize #1}}
\newcommand{\groupname}[1]{\adjustbox{valign=c}{\rotatebox{90}{#1}}}
\newcommand{\predimgssix}[7]{\predfig{#1}{#2} & \predfig{#1}{#3} & \predfig{#1}{#4} & \predfig{#1}{#5} & \predfig{#1}{#6} & \predfig{#1}{#7}}

\begin{tabular}{@{}c@{\hspace{3pt}}l@{}}

    \groupname{Context Frames} &
    \begin{tabular}[c]{@{}ccccccc@{}}
        & $X_{t-9}$ & $X_{t-6}$ & $X_{t-3}$ & $X_{t}$ \\
        \rowname{RGB}   & \rgbctxfig{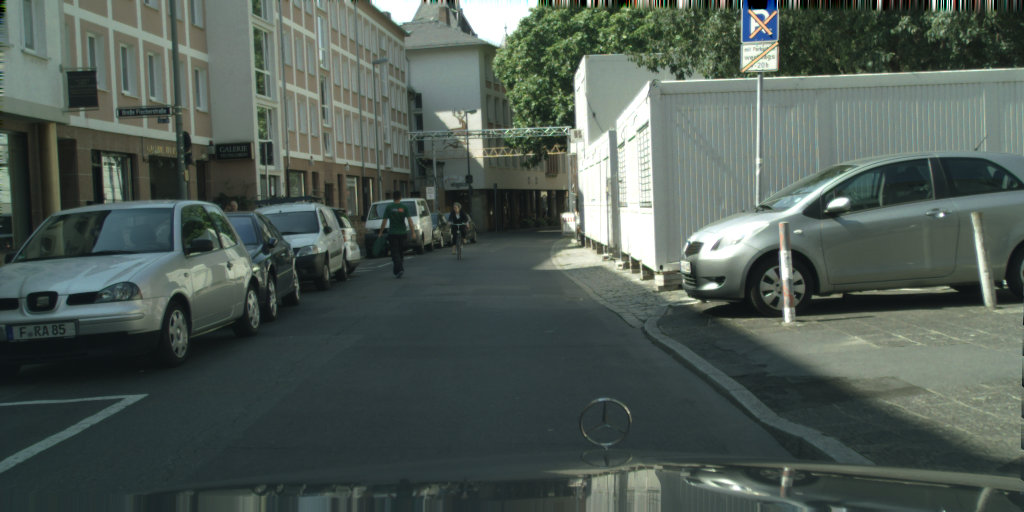} & \rgbctxfig{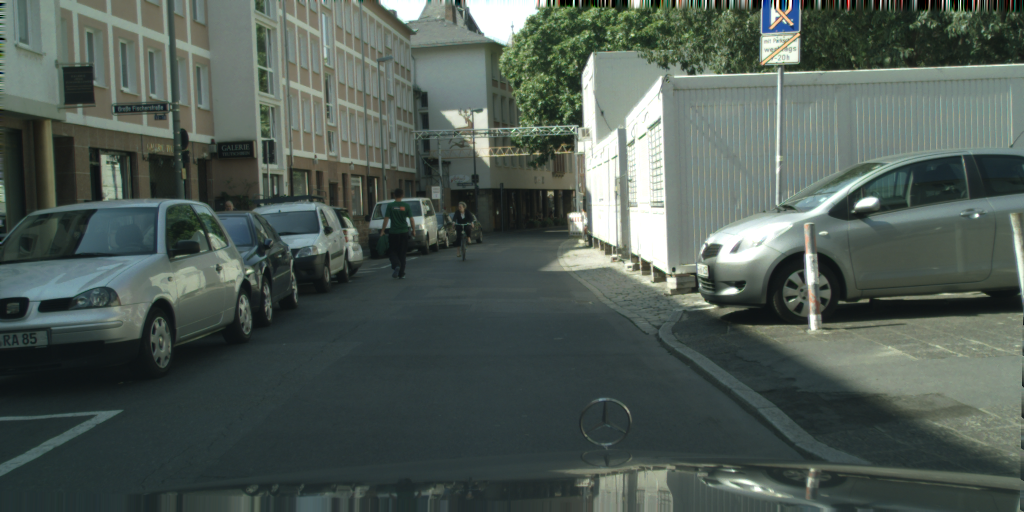} & \rgbctxfig{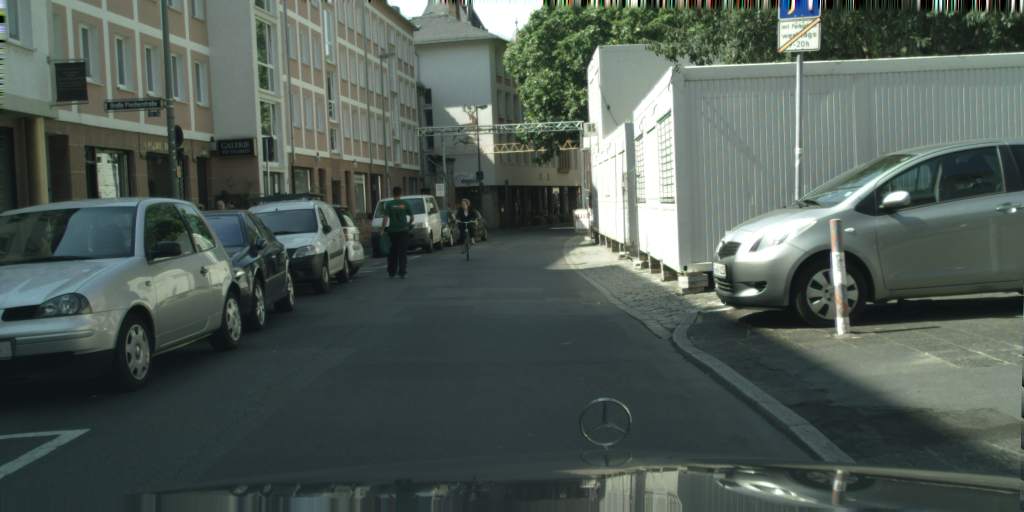} & \rgbctxfig{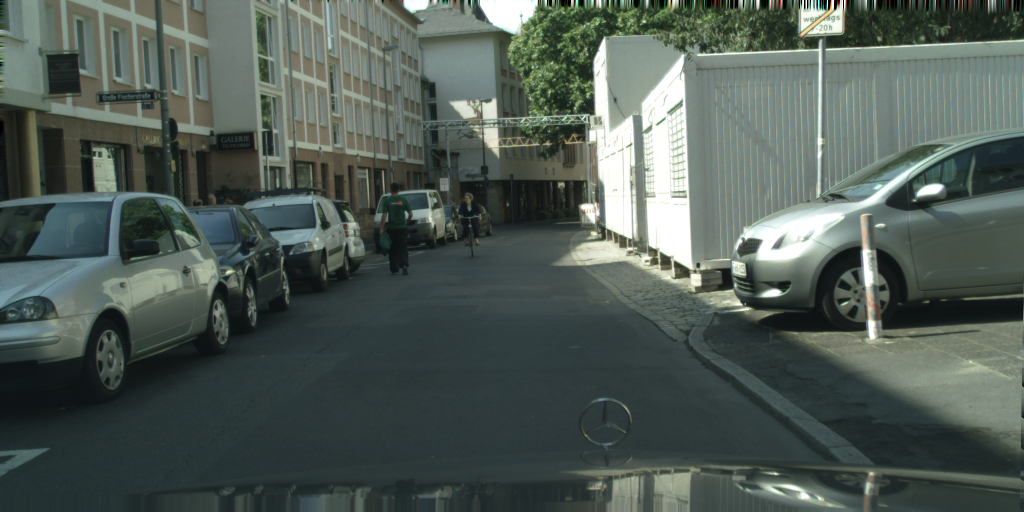} \\[1.2em]
        \rowname{Segm.} & \segmctxfig{t_minus_9} & \segmctxfig{t_minus_6} & \segmctxfig{t_minus_3} & \segmctxfig{t}
    \end{tabular} \\[2ex]

    \groupname{Predicted Frames} &
    \begin{tabular}[c]{@{}ccccccc@{}}
        & $X_{t+9}\,(0.54\text{s})$ & $X_{t+18}\,(1.08\text{s})$ & $X_{t+27}\,(1.62\text{s})$ & $X_{t+36}\,(2.16\text{s})$ & $X_{t+45}\,(2.7\text{s})$ & $X_{t+54}\,(3.24\text{s})$ \\
        \rowname{Dino-\\Foresight}            & \predimgssix{df}{9}{18}{27}{36}{45}{54} \\[1.2em]
        \rowname{Latent-\\Foresight\\(Segm)} & \predimgssix{lf_scale}{9}{18}{27}{36}{45}{54} \\[1.2em]
        \rowname{Latent-\\Foresight\\(Feats)} & \predimgssix{lf_scale_feats}{9}{18}{27}{36}{45}{54}
    \end{tabular} \\

\end{tabular}
}}
    \vspace{-4pt}
    \label{fig:Fig_qual_95}
    \vspace{-5pt}
    \caption{\textbf{Qualitative comparison of long-term forecasting.}
    Given four context frames (top), we forecast up to 3.24s ahead.
    We show semantic segmentation from \texttt{DINO-Foresight} and \ours, and a PCA visualization of \ours predicted features.}
\label{fig:qualitative_long}
\vspace{-20pt}
\end{figure}

\parag{Bottleneck Dimensionality.}
Table~\ref{tab:ablation_dimensionality_compact} examines the effect of the latent bottleneck dimension $C$.
While reconstruction cosine similarity improves monotonically with dimensionality, forecasting performance is strongest at intermediate dimensions ($C \in [128, 512]$).
$C=1152$ achieves the highest reconstruction similarity (only marginally above $C \in [256, 512]$) but substantially degrades mid-term forecasting, highlighting the trade-off between reconstruction fidelity and temporal predictability.
Conversely, excessively small bottlenecks ($C=32$) impair both reconstruction and forecasting.
We select $C=256$, as it provides near-optimal reconstruction and strong forecasting at both horizons (all metrics are reported in \hyperref[tab:ablation_dimensionality]{Appendix Table~\ref*{tab:ablation_dimensionality}}).

\parag{Noise Level Distribution.}
Table~\ref{tab:ablation_noisesched_compact} examines the effect of the noise level distribution $p(t)$.
While reconstruction quality remains largely unchanged, forecasting performance benefits from shifting the logit-normal distribution toward higher noise levels ($\mu=-2$), compared to uniform sampling and the JiT default ($\mu=-0.8$, $\sigma=0.8$).
Increasing its standard deviation to $\sigma=1.5$ further improves mid-term segmentation performance.
We therefore adopt the logit-normal distribution with $(\mu=-2, \sigma=1.5)$ as our default (all metrics are reported in \hyperref[tab:ablation_noisesched]{Appendix Table~\ref*{tab:ablation_noisesched}}).

\parag{Qualitative results.}
Figure~\ref{fig:qualitative_long} compares long-term forecasts of \texttt{DINO-Foresight} and \ours up to 3.24s ahead.
Beyond roughly 2s, \texttt{DINO-Foresight} predictions become nearly static, retaining segmentation artifacts such as spurious regions and noisy segments near the image border.
In contrast, \ours captures the apparent motion of cars and roadside structures under ego-motion, while better preserving scene layout and object boundaries.
Small objects, including pedestrians and cyclists, also remain more clearly delineated across prediction horizons and evolve more plausibly as the ego-vehicle moves.
The PCA visualization of predicted features exhibits similar temporal evolution, suggesting that these dynamics are captured in the predicted feature representations.
Additional qualitative comparisons in \hyperref[sec:additional_viz]{Appendix Subsection~\ref*{sec:additional_viz}}.

\section{Conclusion}

We presented \ours, an end-to-end framework for VFM-based world modeling that jointly learns a feature tokenizer and a flow-based temporal predictor, explicitly shaping latent representations for temporal predictability.
We identified simple but critical design choices that prevent latent collapse and enable stable joint optimization.
Extensive experiments demonstrate consistent improvements over conventional two-stage approaches across multiple future scene understanding tasks and prediction horizons.
Overall, this work establishes end-to-end latent world modeling as a simple and effective alternative to decoupled pipelines, avoiding separate tokenizer and predictor training stages even during high-resolution adaptation.
It also highlights learning representations under temporal prediction objectives as a promising direction for world modeling.

\paragraph{Acknowledgements}

This work has been partially supported by project MIS 5154714 of the National Recovery and Resilience Plan Greece 2.0 funded by the European Union under the NextGenerationEU Program. Hardware resources were granted with the support of GRNET. Also, this work was performed using EuroHPC resources (Project IDS e-dev-2026d01-061 and EHPC-DEV-2026D07-155) and  HPC resources from GENCI-IDRIS (Grants AS011017163 and AD011018152).

\bibliography{iclr2027_conference}

\begin{thebibliography}{58}
\providecommand{\natexlab}[1]{#1}
\providecommand{\url}[1]{\texttt{#1}}
\expandafter\ifx\csname urlstyle\endcsname\relax
  \providecommand{\doi}[1]{doi: #1}\else
  \providecommand{\doi}{doi: \begingroup \urlstyle{rm}\Url}\fi

\bibitem[Ali et~al.(2025)Ali, Bai, Bala, Balaji, Blakeman, Cai, Cao, Cao, Cha, Chao, et~al.]{ali2025world}
Arslan Ali, Junjie Bai, Maciej Bala, Yogesh Balaji, Aaron Blakeman, Tiffany Cai, Jiaxin Cao, Tianshi Cao, Elizabeth Cha, Yu-Wei Chao, et~al.
\newblock World simulation with video foundation models for physical ai.
\newblock \emph{arXiv preprint arXiv:2511.00062}, 2025.

\bibitem[Arai et~al.(2025)Arai, Miwa, Sasaki, Watanabe, Yamaguchi, Aoki, and Yamamoto]{covla}
Hidehisa Arai, Keita Miwa, Kento Sasaki, Kohei Watanabe, Yu~Yamaguchi, Shunsuke Aoki, and Issei Yamamoto.
\newblock Covla: Comprehensive vision-language-action dataset for autonomous driving.
\newblock In \emph{WACV}, 2025.

\bibitem[Assran et~al.(2025)Assran, Bardes, Fan, Garrido, Howes, Muckley, Rizvi, Roberts, Sinha, Zholus, et~al.]{assran2025v}
Mido Assran, Adrien Bardes, David Fan, Quentin Garrido, Russell Howes, Matthew Muckley, Ammar Rizvi, Claire Roberts, Koustuv Sinha, Artem Zholus, et~al.
\newblock V-jepa 2: Self-supervised video models enable understanding, prediction and planning.
\newblock \emph{arXiv preprint arXiv:2506.09985}, 2025.

\bibitem[Ba et~al.(2016)Ba, Kiros, and Hinton]{ba2016layer}
Jimmy~Lei Ba, Jamie~Ryan Kiros, and Geoffrey~E Hinton.
\newblock Layer normalization.
\newblock \emph{arXiv preprint arXiv:1607.06450}, 2016.

\bibitem[Babaeizadeh et~al.(2018)Babaeizadeh, Finn, Erhan, Campbell, and Levine]{babaeizadeh2018stochastic}
Mohammad Babaeizadeh, Chelsea Finn, Dumitru Erhan, Roy~H. Campbell, and Sergey Levine.
\newblock Stochastic variational video prediction.
\newblock In \emph{ICLR}, 2018.
\newblock URL \url{https://openreview.net/forum?id=rk49Mg-CW}.

\bibitem[Baldassarre et~al.(2025)Baldassarre, Szafraniec, Terver, Khalidov, Massa, LeCun, Labatut, Seitzer, and Bojanowski]{baldassarre2025back}
Federico Baldassarre, Marc Szafraniec, Basile Terver, Vasil Khalidov, Francisco Massa, Yann LeCun, Patrick Labatut, Maximilian Seitzer, and Piotr Bojanowski.
\newblock Back to the features: Dino as a foundation for video world models.
\newblock \emph{arXiv preprint arXiv:2507.19468}, 2025.

\bibitem[Balestriero \& LeCun(2025)Balestriero and LeCun]{SigReg}
Randall Balestriero and Yann LeCun.
\newblock Lejepa: Provable and scalable self-supervised learning without the heuristics.
\newblock \emph{arXiv preprint arXiv:2511.08544}, 2025.

\bibitem[Bardes et~al.(2024)Bardes, Garrido, Ponce, Chen, Rabbat, LeCun, Assran, and Ballas]{bardes2024revisiting}
Adrien Bardes, Quentin Garrido, Jean Ponce, Xinlei Chen, Michael Rabbat, Yann LeCun, Mido Assran, and Nicolas Ballas.
\newblock Revisiting feature prediction for learning visual representations from video.
\newblock \emph{Transactions on Machine Learning Research}, 2024.

\bibitem[Boduljak et~al.(2025)Boduljak, Lan, Rupprecht, and Vedaldi]{boduljak2025vfm}
Gabrijel Boduljak, Yushi Lan, Christian Rupprecht, and Andrea Vedaldi.
\newblock Vfmf: World modeling by forecasting vision foundation model features, 2025.
\newblock URL \url{https://arxiv.org/abs/2512.11225}.

\bibitem[Brooks et~al.(2024)Brooks, Peebles, Holmes, DePue, Guo, Jing, Schnurr, Taylor, Luhman, Luhman, et~al.]{videoworldsimulators2024}
Tim Brooks, Bill Peebles, Connor Holmes, Will DePue, Yufei Guo, Li~Jing, David Schnurr, Joe Taylor, Troy Luhman, Eric Luhman, et~al.
\newblock Video generation models as world simulators.
\newblock \emph{OpenAI Blog}, 1:\penalty0 8, 2024.

\bibitem[Caesar et~al.(2020)Caesar, Bankiti, Lang, Vora, Liong, Xu, Krishnan, Pan, Baldan, and Beijbom]{nuscenes}
Holger Caesar, Varun Bankiti, Alex~H. Lang, Sourabh Vora, Venice~Erin Liong, Qiang Xu, Anush Krishnan, Yu~Pan, Giancarlo Baldan, and Oscar Beijbom.
\newblock nuscenes: A multimodal dataset for autonomous driving.
\newblock In \emph{CVPR}, 2020.

\bibitem[Castrejon et~al.(2019)Castrejon, Ballas, and Courville]{castrejon2019improved}
Lluis Castrejon, Nicolas Ballas, and Aaron Courville.
\newblock Improved conditional vrnns for video prediction.
\newblock In \emph{CVPR}, 2019.

\bibitem[Chu et~al.(2026)Chu, Zhang, Han, Li, Yang, Yue, and Guo]{chu2026}
Wenda Chu, Bingliang Zhang, Jiaqi Han, Yizhuo Li, Linjie Yang, Yisong Yue, and Qiushan Guo.
\newblock End-to-end autoregressive image generation with 1d semantic tokenizer, 2026.
\newblock URL \url{https://arxiv.org/abs/2605.00503}.

\bibitem[Cordts et~al.(2016)Cordts, Omran, Ramos, Rehfeld, Enzweiler, Benenson, Franke, Roth, and Schiele]{Cordts_2016_CVPR}
Marius Cordts, Mohamed Omran, Sebastian Ramos, Timo Rehfeld, Markus Enzweiler, Rodrigo Benenson, Uwe Franke, Stefan Roth, and Bernt Schiele.
\newblock The cityscapes dataset for semantic urban scene understanding.
\newblock In \emph{CVPR}, June 2016.

\bibitem[Duggal et~al.(2026)Duggal, Bai, Wu, Zhang, Shechtman, Torralba, Isola, and Freeman]{duggal2026end}
Shivam Duggal, Xingjian Bai, Zongze Wu, Richard Zhang, Eli Shechtman, Antonio Torralba, Phillip Isola, and William~T Freeman.
\newblock End-to-end training for unified tokenization and latent denoising.
\newblock \emph{arXiv preprint arXiv:2603.22283}, 2026.

\bibitem[Gao et~al.(2024)Gao, Yang, Chen, Chitta, Qiu, Geiger, Zhang, and Li]{gao2024vista}
Shenyuan Gao, Jiazhi Yang, Li~Chen, Kashyap Chitta, Yihang Qiu, Andreas Geiger, Jun Zhang, and Hongyang Li.
\newblock Vista: A generalizable driving world model with high fidelity and versatile controllability.
\newblock In \emph{NeurIPS}, 2024.
\newblock URL \url{https://openreview.net/forum?id=Tw9nfNyOMy}.

\bibitem[Gao et~al.(2022)Gao, Tan, Wu, and Li]{gao2022simvp}
Zhangyang Gao, Cheng Tan, Lirong Wu, and Stan~Z Li.
\newblock Simvp: Simpler yet better video prediction.
\newblock In \emph{CVPR}, 2022.

\bibitem[Greff et~al.(2022)Greff, Belletti, Beyer, Doersch, Du, Duckworth, Fleet, Gnanapragasam, Golemo, Herrmann, Kipf, Kundu, Lagun, Laradji, Liu, Meyer, Miao, Nowrouzezahrai, Oztireli, Pot, Radwan, Rebain, Sabour, Sajjadi, Sela, Sitzmann, Stone, Sun, Vora, Wang, Wu, Yi, Zhong, and Tagliasacchi]{Kubric}
Klaus Greff, Francois Belletti, Lucas Beyer, Carl Doersch, Yilun Du, Daniel Duckworth, David~J. Fleet, Dan Gnanapragasam, Florian Golemo, Charles Herrmann, Thomas Kipf, Abhijit Kundu, Dmitry Lagun, Issam Laradji, Hsueh-Ti~(Derek) Liu, Henning Meyer, Yishu Miao, Derek Nowrouzezahrai, Cengiz Oztireli, Etienne Pot, Noha Radwan, Daniel Rebain, Sara Sabour, Mehdi S.~M. Sajjadi, Matan Sela, Vincent Sitzmann, Austin Stone, Deqing Sun, Suhani Vora, Ziyu Wang, Tianhao Wu, Kwang~Moo Yi, Fangcheng Zhong, and Andrea Tagliasacchi.
\newblock Kubric: A scalable dataset generator.
\newblock In \emph{CVPR}, 2022.

\bibitem[Gupta et~al.(2023)Gupta, Tian, Zhang, Wu, Mart{\'\i}n-Mart{\'\i}n, and Fei-Fei]{gupta2023maskvit}
Agrim Gupta, Stephen Tian, Yunzhi Zhang, Jiajun Wu, Roberto Mart{\'\i}n-Mart{\'\i}n, and Li~Fei-Fei.
\newblock Maskvit: Masked visual pre-training for video prediction.
\newblock In \emph{ICLR}, 2023.
\newblock URL \url{https://openreview.net/forum?id=QAV2CcLEDh}.

\bibitem[Harvey et~al.(2022)Harvey, Naderiparizi, Masrani, Weilbach, and Wood]{harvey2022flexible}
William Harvey, Saeid Naderiparizi, Vaden Masrani, Christian~Dietrich Weilbach, and Frank Wood.
\newblock Flexible diffusion modeling of long videos.
\newblock In \emph{NeurIPS}, 2022.
\newblock URL \url{https://openreview.net/forum?id=0RTJcuvHtIu}.

\bibitem[Henry et~al.(2020)Henry, Dachapally, Pawar, and Chen]{henry2020query}
Alex Henry, Prudhvi~Raj Dachapally, Shubham~Shantaram Pawar, and Yuxuan Chen.
\newblock Query-key normalization for transformers.
\newblock In \emph{Findings of the Association for Computational Linguistics: EMNLP 2020}, pp.\  4246--4253, 2020.

\bibitem[Ho et~al.(2022{\natexlab{a}})Ho, Chan, Saharia, Whang, Gao, Gritsenko, Kingma, Poole, Norouzi, Fleet, et~al.]{ho2022imagen}
Jonathan Ho, William Chan, Chitwan Saharia, Jay Whang, Ruiqi Gao, Alexey Gritsenko, Diederik~P Kingma, Ben Poole, Mohammad Norouzi, David~J Fleet, et~al.
\newblock Imagen video: High definition video generation with diffusion models.
\newblock \emph{arXiv preprint arXiv:2210.02303}, 2022{\natexlab{a}}.

\bibitem[Ho et~al.(2022{\natexlab{b}})Ho, Salimans, Gritsenko, Chan, Norouzi, and Fleet]{ho2022video}
Jonathan Ho, Tim Salimans, Alexey Gritsenko, William Chan, Mohammad Norouzi, and David~J Fleet.
\newblock Video diffusion models.
\newblock In \emph{NeurIPS}, 2022{\natexlab{b}}.
\newblock URL \url{https://proceedings.neurips.cc/paper_files/paper/2022/file/39235c56aef13fb05a6adc95eb9d8d66-Paper-Conference.pdf}.

\bibitem[Ioffe \& Szegedy(2015)Ioffe and Szegedy]{ioffe2015batch}
Sergey Ioffe and Christian Szegedy.
\newblock Batch normalization: Accelerating deep network training by reducing internal covariate shift.
\newblock In \emph{ICML}, 2015.

\bibitem[Karypidis et~al.(2025{\natexlab{a}})Karypidis, Kakogeorgiou, Gidaris, and Komodakis]{karypidis2025advancing}
Efstathios Karypidis, Ioannis Kakogeorgiou, Spyros Gidaris, and Nikos Komodakis.
\newblock Advancing semantic future prediction through multimodal visual sequence transformers.
\newblock In \emph{CVPR}, 2025{\natexlab{a}}.

\bibitem[Karypidis et~al.(2025{\natexlab{b}})Karypidis, Kakogeorgiou, Gidaris, and Komodakis]{karypidis2025dinoforesight}
Efstathios Karypidis, Ioannis Kakogeorgiou, Spyros Gidaris, and Nikos Komodakis.
\newblock {DINO}-foresight: Looking into the future with {DINO}.
\newblock In \emph{NeurIPS}, 2025{\natexlab{b}}.
\newblock URL \url{https://openreview.net/forum?id=gimtybo07H}.

\bibitem[Karypidis et~al.(2026)Karypidis, Gidaris, and Komodakis]{karypidisre2pix}
Efstathios Karypidis, Spyros Gidaris, and Nikos Komodakis.
\newblock Representations before pixels: Semantics-guided hierarchical video prediction.
\newblock In \emph{ECCV}, 2026.

\bibitem[Kerssies et~al.(2026)Kerssies, Berton, He, Yu, Ma, de~Geus, Dubbelman, and Chen]{kerssies2026frame}
Tommie Kerssies, Gabriele Berton, Ju~He, Qihang Yu, Wufei Ma, Daan de~Geus, Gijs Dubbelman, and Liang-Chieh Chen.
\newblock A frame is worth one token: Efficient generative world modeling with delta tokens.
\newblock \emph{CVPR}, 2026.

\bibitem[Kingma \& Welling(2013)Kingma and Welling]{kingma2013auto}
Diederik~P Kingma and Max Welling.
\newblock Auto-encoding variational bayes.
\newblock \emph{arXiv preprint arXiv:1312.6114}, 2013.

\bibitem[Lee et~al.(2018)Lee, Zhang, Ebert, Abbeel, Finn, and Levine]{lee2018stochastic}
Alex~X Lee, Richard Zhang, Frederik Ebert, Pieter Abbeel, Chelsea Finn, and Sergey Levine.
\newblock Stochastic adversarial video prediction.
\newblock \emph{arXiv preprint arXiv:1804.01523}, 2018.

\bibitem[Lee et~al.(2021)Lee, Kim, Choi, Kim, and Ro]{lee2021video}
Sangmin Lee, Hak~Gu Kim, Dae~Hwi Choi, Hyung-Il Kim, and Yong~Man Ro.
\newblock Video prediction recalling long-term motion context via memory alignment learning.
\newblock In \emph{CVPR}, 2021.

\bibitem[Leng et~al.(2025)Leng, Singh, Hou, Xing, Xie, and Zheng]{leng2025repa}
Xingjian Leng, Jaskirat Singh, Yunzhong Hou, Zhenchang Xing, Saining Xie, and Liang Zheng.
\newblock Repa-e: Unlocking vae for end-to-end tuning of latent diffusion transformers.
\newblock In \emph{CVPR}, 2025.

\bibitem[Li \& He(2025)Li and He]{li2025back}
Tianhong Li and Kaiming He.
\newblock Back to basics: Let denoising generative models denoise.
\newblock \emph{arXiv preprint arXiv:2511.13720}, 2025.

\bibitem[Lin et~al.(2021)Lin, Sun, Hu, Yu, Lai, and Zheng]{lin2021predictive}
Zihang Lin, Jiangxin Sun, Jian-Fang Hu, Qizhi Yu, Jian-Huang Lai, and Wei-Shi Zheng.
\newblock Predictive feature learning for future segmentation prediction.
\newblock In \emph{CVPR}, 2021.

\bibitem[Loshchilov \& Hutter(2019)Loshchilov and Hutter]{loshchilov2018decoupled}
Ilya Loshchilov and Frank Hutter.
\newblock Decoupled weight decay regularization.
\newblock In \emph{ICLR}, 2019.
\newblock URL \url{https://openreview.net/forum?id=Bkg6RiCqY7}.

\bibitem[Luc et~al.(2017)Luc, Neverova, Couprie, Verbeek, and LeCun]{luc2017predicting}
Pauline Luc, Natalia Neverova, Camille Couprie, Jakob Verbeek, and Yann LeCun.
\newblock Predicting deeper into the future of semantic segmentation.
\newblock In \emph{ICCV}, 2017.

\bibitem[Maes et~al.(2026)Maes, Lidec, Scieur, LeCun, and Balestriero]{maes2026leworldmodel}
Lucas Maes, Quentin~Le Lidec, Damien Scieur, Yann LeCun, and Randall Balestriero.
\newblock Leworldmodel: Stable end-to-end joint-embedding predictive architecture from pixels.
\newblock \emph{arXiv preprint arXiv:2603.19312}, 2026.

\bibitem[Nabavi et~al.(2018)Nabavi, Rochan, and Wang]{nabavi2018future}
Seyed~Shahabeddin Nabavi, Mrigank Rochan, and Yang Wang.
\newblock Future semantic segmentation with convolutional lstm.
\newblock In \emph{BMVC}, 2018.

\bibitem[Oquab et~al.(2024)Oquab, Darcet, Moutakanni, Vo, Szafraniec, Khalidov, Fernandez, HAZIZA, Massa, El-Nouby, Assran, Ballas, Galuba, Howes, Huang, Li, Misra, Rabbat, Sharma, Synnaeve, Xu, Jegou, Mairal, Labatut, Joulin, and Bojanowski]{oquab2024dinov}
Maxime Oquab, Timoth{\'e}e Darcet, Th{\'e}o Moutakanni, Huy~V. Vo, Marc Szafraniec, Vasil Khalidov, Pierre Fernandez, Daniel HAZIZA, Francisco Massa, Alaaeldin El-Nouby, Mido Assran, Nicolas Ballas, Wojciech Galuba, Russell Howes, Po-Yao Huang, Shang-Wen Li, Ishan Misra, Michael Rabbat, Vasu Sharma, Gabriel Synnaeve, Hu~Xu, Herve Jegou, Julien Mairal, Patrick Labatut, Armand Joulin, and Piotr Bojanowski.
\newblock {DINO}v2: Learning robust visual features without supervision.
\newblock \emph{Transactions on Machine Learning Research}, 2024.
\newblock URL \url{https://openreview.net/forum?id=a68SUt6zFt}.

\bibitem[Pan et~al.(2026)Pan, Feng, Dai, Wang, Lin, Guo, Luo, and Zheng]{SemVAE}
Yueming Pan, Ruoyu Feng, Qi~Dai, Yuqi Wang, Wenfeng Lin, Mingyu Guo, Chong Luo, and Nanning Zheng.
\newblock Semantics lead the way: Harmonizing semantic and texture modeling with asynchronous latent diffusion.
\newblock In \emph{CVPR}, 2026.

\bibitem[Peebles \& Xie(2023)Peebles and Xie]{peebles2023scalable}
William Peebles and Saining Xie.
\newblock Scalable diffusion models with transformers.
\newblock In \emph{ICCV}, 2023.

\bibitem[Porcher et~al.(2026)Porcher, Carion, Alahari, and Chen]{porcher2026flow}
Francois Porcher, Nicolas Carion, Karteek Alahari, and Shizhe Chen.
\newblock Flow matching in feature space for stochastic world modeling.
\newblock \emph{arXiv preprint arXiv:2606.29059}, 2026.

\bibitem[Ranftl et~al.(2021)Ranftl, Bochkovskiy, and Koltun]{ranftl2021vision}
Ren{\'e} Ranftl, Alexey Bochkovskiy, and Vladlen Koltun.
\newblock Vision transformers for dense prediction.
\newblock In \emph{CVPR}, 2021.

\bibitem[Shazeer(2020)]{shazeer2020glu}
Noam Shazeer.
\newblock Glu variants improve transformer.
\newblock \emph{arXiv preprint arXiv:2002.05202}, 2020.

\bibitem[Sun et~al.(2026)Sun, Wang, Zhang, Liu, Jia, Song, Huang, and Luo]{sun2026vggt}
Xiangyu Sun, Shijie Wang, Fengyi Zhang, Lin Liu, Caiyan Jia, Ziying Song, Zi~Huang, and Yadan Luo.
\newblock Vggt-world: Transforming vggt into an autoregressive geometry world model.
\newblock \emph{arXiv preprint arXiv:2603.12655}, 2026.

\bibitem[Vondrick et~al.(2016)Vondrick, Pirsiavash, and Torralba]{vondrick2016anticipating}
Carl Vondrick, Hamed Pirsiavash, and Antonio Torralba.
\newblock Anticipating visual representations from unlabeled video.
\newblock In \emph{CVPR}, 2016.

\bibitem[Walker et~al.(2025)Walker, V{\'e}lez, Cabrera, Zhou, Kabra, Doersch, Ovsjanikov, Carreira, and Ginosar]{walker2025generalist}
Jacob~C Walker, Pedro V{\'e}lez, Luisa~Polania Cabrera, Guangyao Zhou, Rishabh Kabra, Carl Doersch, Maks Ovsjanikov, Jo{\~a}o Carreira, and Shiry Ginosar.
\newblock Generalist forecasting with frozen video models via latent diffusion.
\newblock \emph{arXiv preprint arXiv:2507.13942}, 2025.

\bibitem[Wang et~al.(2024)Wang, Zhang, Luo, Sun, Cui, Wang, Zhang, Wang, Li, Yu, et~al.]{wang2024emu3}
Xinlong Wang, Xiaosong Zhang, Zhengxiong Luo, Quan Sun, Yufeng Cui, Jinsheng Wang, Fan Zhang, Yueze Wang, Zhen Li, Qiying Yu, et~al.
\newblock Emu3: Next-token prediction is all you need.
\newblock \emph{arXiv preprint arXiv:2409.18869}, 2024.

\bibitem[Wang et~al.(2018)Wang, Jiang, Yang, Li, Long, and Fei-Fei]{wang2018eidetic}
Yunbo Wang, Lu~Jiang, Ming-Hsuan Yang, Li-Jia Li, Mingsheng Long, and Li~Fei-Fei.
\newblock Eidetic 3d lstm: A model for video prediction and beyond.
\newblock In \emph{ICLR}, 2018.

\bibitem[Wu et~al.(2021)Wu, Yao, Wang, and Long]{wu2021motionrnn}
Haixu Wu, Zhiyu Yao, Jianmin Wang, and Mingsheng Long.
\newblock Motionrnn: A flexible model for video prediction with spacetime-varying motions.
\newblock In \emph{CVPR}, 2021.

\bibitem[Xu et~al.(2018)Xu, Ni, Li, Cheng, and Yang]{Xu_2018_CVPR}
Jingwei Xu, Bingbing Ni, Zefan Li, Shuo Cheng, and Xiaokang Yang.
\newblock Structure preserving video prediction.
\newblock In \emph{CVPR}, June 2018.

\bibitem[Yan et~al.(2021)Yan, Zhang, Abbeel, and Srinivas]{yan2021videogpt}
Wilson Yan, Yunzhi Zhang, Pieter Abbeel, and Aravind Srinivas.
\newblock Videogpt: Video generation using vq-vae and transformers.
\newblock \emph{arXiv preprint arXiv:2104.10157}, 2021.

\bibitem[Yang et~al.(2024{\natexlab{a}})Yang, Kang, Huang, Xu, Feng, and Zhao]{yang_depth}
Lihe Yang, Bingyi Kang, Zilong Huang, Xiaogang Xu, Jiashi Feng, and Hengshuang Zhao.
\newblock Depth anything: Unleashing the power of large-scale unlabeled data.
\newblock In \emph{CVPR}, 2024{\natexlab{a}}.

\bibitem[Yang et~al.(2024{\natexlab{b}})Yang, Kang, Huang, Zhao, Xu, Feng, and Zhao]{yang2024depth}
Lihe Yang, Bingyi Kang, Zilong Huang, Zhen Zhao, Xiaogang Xu, Jiashi Feng, and Hengshuang Zhao.
\newblock Depth anything v2.
\newblock In \emph{NeurIPS}, 2024{\natexlab{b}}.
\newblock URL \url{https://openreview.net/forum?id=cFTi3gLJ1X}.

\bibitem[Yu et~al.(2023)Yu, Cheng, Sohn, Lezama, Zhang, Chang, Hauptmann, Yang, Hao, Essa, et~al.]{yu2023magvit}
Lijun Yu, Yong Cheng, Kihyuk Sohn, Jos{\'e} Lezama, Han Zhang, Huiwen Chang, Alexander~G Hauptmann, Ming-Hsuan Yang, Yuan Hao, Irfan Essa, et~al.
\newblock Magvit: Masked generative video transformer.
\newblock In \emph{CVPR}, 2023.

\bibitem[Yu et~al.(2024)Yu, Lezama, Gundavarapu, Versari, Sohn, Minnen, Cheng, Gupta, Gu, Hauptmann, Gong, Yang, Essa, Ross, and Jiang]{yu2024language}
Lijun Yu, Jose Lezama, Nitesh~Bharadwaj Gundavarapu, Luca Versari, Kihyuk Sohn, David Minnen, Yong Cheng, Agrim Gupta, Xiuye Gu, Alexander~G Hauptmann, Boqing Gong, Ming-Hsuan Yang, Irfan Essa, David~A Ross, and Lu~Jiang.
\newblock Language model beats diffusion - tokenizer is key to visual generation.
\newblock In \emph{ICLR}, 2024.
\newblock URL \url{https://openreview.net/forum?id=gzqrANCF4g}.

\bibitem[Zhang \& Sennrich(2019)Zhang and Sennrich]{rmsnorm}
Biao Zhang and Rico Sennrich.
\newblock Root mean square layer normalization.
\newblock In \emph{NeurIPS}, 2019.
\newblock URL \url{https://proceedings.neurips.cc/paper_files/paper/2019/file/1e8a19426224ca89e83cef47f1e7f53b-Paper.pdf}.

\bibitem[Zhou et~al.(2025)Zhou, Pan, LeCun, and Pinto]{zhou2025dinowm}
Gaoyue Zhou, Hengkai Pan, Yann LeCun, and Lerrel Pinto.
\newblock {DINO}-{WM}: World models on pre-trained visual features enable zero-shot planning.
\newblock In \emph{ICML}, 2025.
\newblock URL \url{https://openreview.net/forum?id=D5RNACOZEI}.

\end{thebibliography}
\bibliographystyle{iclr2027_conference}

\appendix
\newpage
\section{Appendix}

\subsection{Additional Results.}\label{sec:appendix-results}  

\paragraph{Full Comparison on Cityscapes.}
Table~\ref{tab:ablation_sota_full} extends Table~\ref{tab:comparison_with_prior_work} with all metrics, including depth AbsRel and surface normal mean angular error, and adds a Copy Last baseline that repeats the last context frame.
The sharp drop of Copy Last at the mid-term horizon confirms that the benchmark requires modeling non-trivial scene dynamics.
\ours outperforms \texttt{DINO-Foresight} on semantic segmentation, depth $\delta_1$, and both surface normal metrics at both horizons, with the largest gains at the mid-term horizon.
For depth AbsRel, both methods perform comparably.

\paragraph{Reconstruction Loss.}
Table~\ref{tab:ablation_reconloss} compares reconstruction objectives for the autoencoder.
Using only the cosine similarity term fails to preserve feature information, substantially degrading both reconstruction and forecasting, likely because the scale-invariant cosine loss leaves feature magnitudes unconstrained.
MSE alone and the combination of MSE and cosine similarity perform comparably; we adopt the combined objective following~\citet{SemVAE}.

\paragraph{Bottleneck Dimensionality and Noise Level Distribution.}
Tables~\ref{tab:ablation_dimensionality} and~\ref{tab:ablation_noisesched} report all metrics for the ablations in Tables~\ref{tab:ablation_dimensionality_compact} and~\ref{tab:ablation_noisesched_compact}.
Segmentation over all classes (ALL) and predicted-feature cosine similarity follow the same trends as the movable-object mIoU discussed in Section~\ref{sec:experiments}: forecasting is strongest at intermediate bottleneck dimensions and degrades at $C=1152$, while shifting the logit-normal distribution to $\mu=-2$ improves forecasting over uniform sampling and the JiT default, with $\sigma=1.5$ yielding the best mid-term performance.

\begin{table*}[!ht]
\footnotesize
\centering
\caption{\textbf{Comparison across tasks on Cityscapes}. 
We use DINOv2 encoder and show performance on segmentation (ALL, MO), depth estimation ($\delta_{1}$ accuracy, AbsRel error), and surface normal prediction (m, percentage within 11.25°). \texttt{VISTA$_{ft}$} is the \texttt{VISTA} model fine-tuned on Cityscapes.} 
\resizebox{\columnwidth}{!}{
\begin{tabular}{l C{0.75cm}C{0.75cm}C{0.75cm}C{0.75cm} @{\hspace{1.5em}}C{0.75cm}C{0.75cm}C{0.75cm}C{0.75cm} C{0.75cm}C{0.75cm}C{0.75cm}C{0.75cm}}
\toprule
 \multirow{3}{*}[-1.0ex]{\Th{Method}} & \multicolumn{4}{c}{\Th{Semantic Segm.}} & \multicolumn{4}{c}{\Th{Depth}} & \multicolumn{4}{c}{\Th{Surface Normals}} \\
\cmidrule(lr){2-5} \cmidrule(lr){6-9} \cmidrule(lr){10-13}
& \multicolumn{2}{c}{\Th{Short}} & \multicolumn{2}{c}{\Th{Mid}} & \multicolumn{2}{c}{\Th{Short}} & \multicolumn{2}{c}{\Th{Mid}} & \multicolumn{2}{c}{\Th{Short}} & \multicolumn{2}{c}{\Th{Mid}} \\
\cmidrule(r){2-3} \cmidrule(lr){4-5} \cmidrule(lr){6-7} \cmidrule(lr){8-9} \cmidrule(lr){10-11} \cmidrule(lr){12-13}
& ALL$\uparrow$ & MO$\uparrow$ & ALL$\uparrow$ & MO$\uparrow$ & $\delta_{1}$ $\uparrow$ & AbsR$\downarrow$ & $\delta_{1}$$\uparrow$ & AbsR$\downarrow$ & m$\downarrow$ & 11.25$^\circ$$\uparrow$ & m$\downarrow$ & 11.25$^\circ$$\uparrow$ \\
\midrule
\graycell{\texttt{Oracle}} & \graycell{77.1} & \graycell{77.3} & \graycell{77.1} & \graycell{77.3} & \graycell{89.6} & \graycell{.103} & \graycell{89.6} & \graycell{.103} & \graycell{2.88} & \graycell{96.3} & \graycell{2.88} & \graycell{96.3} \\
\texttt{Copy Last} & 54.7 & 52.0 & 40.4 & 32.3 & 84.1 & .154 & 77.8 & .212 & 4.41 & 89.2 & 5.39 & 84.0 \\
\Th{\texttt{VISTA$_{ft}$}}& 64.9 & 62.1 & 53.9 & 51.0 & 86.4 & .124 & 82.8 & .153 & 3.75 & 93.0 & 4.30 & 90.0 \\
\texttt{Dino-Foresight} & 71.8 & 71.7 & 59.8 & 57.6 & 88.6 &  \textbf{.114} & 85.4 &  \textbf{.136} & 3.39 & 94.4 & 4.00 & 91.3 \\
\rowcolor{baselinecolor} \ours  & \underline{72.8} & \underline{72.6} & \underline{61.7} & \underline{59.9} & \underline{88.9} & \underline{.118} & \underline{86.6} & \underline{.137} & \underline{3.13} & \underline{95}  & \underline{3.77}  & \underline{92} \\
\rowcolor{baselinecolor} \oursplus  &  \textbf{73.3} &  \textbf{72.9} &  \textbf{63.3} & \textbf{61.9} & \textbf{88.9} & .120 & \textbf{86.9} & \textbf{.136} & \textbf{3.11} & \textbf{95.2} & \textbf{3.68} & \textbf{92.4} \\

\bottomrule
\end{tabular}
}
\label{tab:ablation_sota_full}
\end{table*}
\vspace{-10pt}
\begin{table}[ht]
\caption{\textbf{Ablation of Reconstruction Loss.} We vary the reconstruction objective used to train the autoencoder and report reconstruction fidelity and prediction quality at short- and mid-term horizons. All results are reported at $224 \times 448$ resolution.}
\scriptsize
\centering
\setlength{\tabcolsep}{1pt}
\begin{tabular}{l @{\hspace{1.5em}} C{1.4cm} C{0.75cm} C{0.75cm} @{\hspace{1.5em}} C{1.4cm} C{0.75cm} C{0.75cm} @{\hspace{1.5em}} C{1.4cm} C{0.75cm} C{0.75cm}}
\toprule
\multirow{2}{*}[-0.5ex]{\Th{Method}}
& \multicolumn{3}{c}{\Th{Reconstruction}}
& \multicolumn{3}{c}{\Th{Prediction (Short-Term)}}
& \multicolumn{3}{c}{\Th{Prediction (Mid-Term)}} \\
\cmidrule(lr){2-4} \cmidrule(lr){5-7} \cmidrule(lr){8-10}
& Cos Sim$\uparrow$ & ALL$\uparrow$ & MO$\uparrow$
& Cos Sim$\uparrow$ & ALL$\uparrow$ & MO$\uparrow$
& Cos Sim$\uparrow$ & ALL$\uparrow$ & MO$\uparrow$ \\
\midrule
Cosine Only & 0.889 & 27.50 & 15.78 & 0.895 & 33.22 & 24.23 & 0.881 & 32.46 & 22.92 \\
MSE Only    & 0.988 & 68.26 & 67.28 & 0.971 & 65.36 & 64.14 & 0.943 & 56.54 & 53.59 \\
\rowcolor{baselinecolor} MSE + Cosine & 0.989 & 68.22 & 67.26 & 0.971 & 65.24 & 64.37 & 0.944 & 56.60 & 53.42 \\
\bottomrule
\end{tabular}
\label{tab:ablation_reconloss}
\end{table}
\begin{table}[t]
\caption{\textbf{Ablation of Bottleneck Dimensionality.} We vary the latent dimension $C$ of the autoencoder and report reconstruction fidelity and prediction quality at short- and mid-term horizons. All results are reported at $224 \times 448$ resolution and do not include RoPE in the autoencoder.}
\scriptsize
\centering
\setlength{\tabcolsep}{1pt}
\begin{tabular}{l @{\hspace{1.5em}} C{1.4cm} C{0.75cm} C{0.75cm} @{\hspace{1.5em}} C{1.4cm} C{0.75cm} C{0.75cm} @{\hspace{1.5em}} C{1.4cm} C{0.75cm} C{0.75cm}}
\toprule
\multirow{2}{*}[-0.5ex]{\Th{Dimensionality}}
& \multicolumn{3}{c}{\Th{Reconstruction}}
& \multicolumn{3}{c}{\Th{Prediction (Short-Term)}}
& \multicolumn{3}{c}{\Th{Prediction (Mid-Term)}} \\
\cmidrule(lr){2-4} \cmidrule(lr){5-7} \cmidrule(lr){8-10}
& Cos Sim$\uparrow$ & ALL$\uparrow$ & MO$\uparrow$
& Cos Sim$\uparrow$ & ALL$\uparrow$ & MO$\uparrow$
& Cos Sim$\uparrow$ & ALL$\uparrow$ & MO$\uparrow$ \\
\midrule
32  & 0.980 & 64.90 & 61.84 & 0.966 & 62.59 & 59.93 & 0.941 & 54.37 & 49.82 \\
64  & 0.984 & 67.14 & 65.79 & 0.969 & 64.24 & 62.48 & 0.943 & 56.57 & 54.03 \\
128 & 0.986 & 67.77 & 66.51 & 0.971 & 65.26 & 64.03 & 0.945 & 56.95 & 54.12\\
\rowcolor{baselinecolor} 256 & 0.989 &  68.22 & 67.26 & 0.971 & 65.42 & 64.37 & 0.944 & 56.60 & 53.42 \\
512  & 0.990 & 68.21 & 67.18 & 0.971 & 65.23 & 64.00 & 0.942 & 56.80 & 54.24 \\
1152  & 0.991 & 68.08 & 66.94 & 0.970 & 64.95 & 63.98 & 0.936 & 55.07 & 51.99\\
\bottomrule
\end{tabular}
\label{tab:ablation_dimensionality}
\end{table}
\begin{table}[t]
\caption{\textbf{Ablation of Noise Level Distribution $p(t)$.} We compare uniform sampling and logit-normal distributions with different mean and standard deviation parameters for training noise levels. All results are reported at $224 \times 448$ resolution, do not include RoPE in the autoencoder, and use a latent dimension of $C=512$.}
\scriptsize
\centering
\setlength{\tabcolsep}{2pt}
\begin{tabular}{l @{\hspace{1.5em}} C{1.4cm} C{0.75cm} C{0.75cm} @{\hspace{1.5em}} C{1.4cm} C{0.75cm} C{0.75cm} @{\hspace{1.5em}} C{1.4cm} C{0.75cm} C{0.75cm}}
\toprule
\multirow{2}{*}[-0.5ex]{\Th{Noise Distribution}}
& \multicolumn{3}{c}{\Th{Reconstruction}}
& \multicolumn{3}{c}{\Th{Prediction (Short-Term)}}
& \multicolumn{3}{c}{\Th{Prediction (Mid-Term)}} \\
\cmidrule(lr){2-4} \cmidrule(lr){5-7} \cmidrule(lr){8-10}
& Cos Sim$\uparrow$ & ALL$\uparrow$ & MO$\uparrow$
& Cos Sim$\uparrow$ & ALL$\uparrow$ & MO$\uparrow$
& Cos Sim$\uparrow$ & ALL$\uparrow$ & MO$\uparrow$ \\
\midrule
Uniform  & 0.990 & 68.17 & 67.17 & 0.969 & 64.34 & 63.11 & 0.937 & 54.60 & 51.00 \\
Logit-Normal (-0.8,0.8)  & 0.990 & 68.18 & 67.18 & 0.967 & 63.73 & 62.75 & 0.933 & 53.94 & 50.89 \\
Logit-Normal (-2,0.8)  & 0.990 & 68.22 & 67.18 & 0.971 & 65.35 & 64.40 & 0.941 & 56.30 & 53.42 \\
\rowcolor{baselinecolor} Logit-Normal (-2,1.5) & 0.990 & 68.21 & 67.18 & 0.971 & 65.23 & 64.00 & 0.942 & 56.80 & 54.24 \\

\bottomrule
\end{tabular}
\label{tab:ablation_noisesched}
\end{table}

\subsection{Datasets.}
\label{app:datasets}
\textbf{Cityscapes}~\citep{Cordts_2016_CVPR} provides 2,975 training and 500 validation video sequences captured at 16 fps with a resolution of $1024 \times 2048$ pixels. Each sequence consists of 30 frames, with the 20th frame annotated for semantic segmentation across 19 classes.
\textbf{nuScenes}~\citep{nuscenes} comprises 700 training and 150 validation scenes recorded at 12 Hz, with each scene spanning 20 seconds of urban driving.
\textbf{Kubric}~\citep{Kubric} is a synthetic benchmark of multi-object scenes with controlled dynamics, which we use to evaluate generalization beyond the driving domain. We use the MOVi-A configuration, comprising 9,703 training and 250 validation sequences captured at 12 fps with a resolution of $256 \times 256$ pixels. Each sequence consists of 24 frames depicting 3--10 rigid objects moving on a static background with collisions, with full per-frame annotations (segmentation, depth, flow, 3D) available.
\textbf{CoVLA}~\citep{covla} is a real-world driving dataset comprising 10,000 video sequences spanning over 80 hours, captured at 20 Hz, with accompanying vehicle state, trajectory, and scene annotations. We use CoVLA alongside Cityscapes and nuScenes to train \oursplus, demonstrating that our approach benefits from larger and more diverse driving data.

\subsection{Implementation Details}
\label{sec:appendix-implemdetails}

\paragraph{Autoencoder and Predictor Architecture.}
The autoencoder $\mathcal{E}_\phi, \mathcal{D}_\psi$ consists of a transformer encoder and decoder, each with 2 layers, a hidden dimension of $1536$, and $6$ attention heads. The predictor operates on the latent space defined by the autoencoder and consists of 12 layers with a hidden dimension of $d = 1152$, processing sequences of $N = 5$ frames ($N_c = 4$ context frames and $N_p = 1$ future frame). The noise level $t$ is injected via an adaLN-zero conditioning module, which produces a shared set of modulation parameters applied across all transformer layers.

\paragraph{Optimization and Training.}
For end-to-end training, we use the AdamW optimizer~\citep{loshchilov2018decoupled} with momentum parameters $\beta_1 = 0.9$, $\beta_2 = 0.99$, weight decay $0.2$, and a learning rate of $6.4 \times 10^{-4}$ with cosine annealing. Following~\citet{karypidis2025dinoforesight}, for driving datasets we train at low resolution and fine-tune at high resolution: models are pretrained at $224\times448$ and subsequently fine-tuned at $448\times896$ for evaluation against prior work. For Kubric, we train only at $224\times224$ resolution for 800 epochs. Training is conducted on 8 GH200 GPUs with an effective batch size of 64.

\paragraph{Downstream heads.} We provide implementation details for the heads trained on different downstream tasks. We train frozen DPT heads~\citep{ranftl2021vision} for semantic segmentation, depth estimation, and surface normal prediction, using the implementation from Depth Anything~\citep{yang_depth, yang2024depth} with feature dimensionality 256 and \texttt{dpt\_out\_channels = [128, 256, 512, 512]}. All heads are trained for 100 epochs with a batch size of 128 on $16 \times 8$ GPUs, using AdamW with a learning rate of $1.6 \times 10^{-3}$, linear warmup for the first 10 epochs, and weight decay $10^{-4}$. For semantic segmentation, we use a polynomial learning rate schedule and cross-entropy loss over 19 classes. For depth estimation, we use cosine annealing and cross-entropy loss over 256 discretized depth bins. For surface normal estimation, we use a polynomial schedule and a combined cosine similarity and $L_2$ loss with weighted averaging.

\subsection{Additional Visualizations}
\label{sec:additional_viz}
\begin{figure}[h]
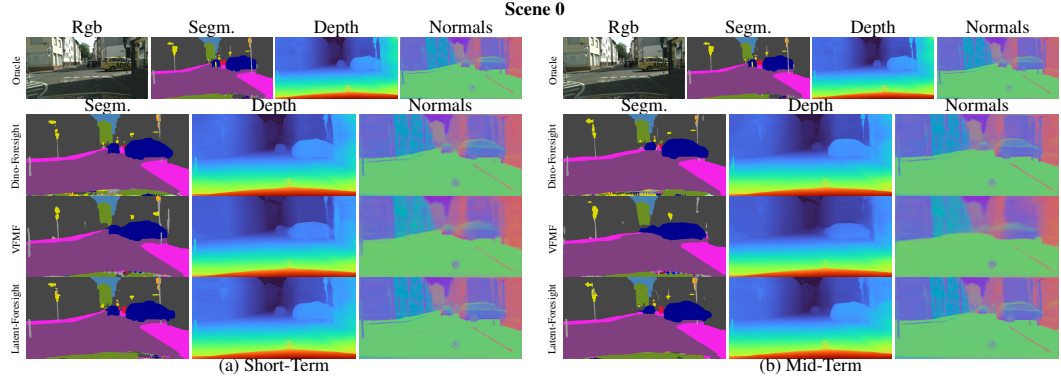

    \centering
    \scriptsize
\centering

\newcommand{\lfscene}{figs/scene_0}
\newcommand{\lfscenenum}{0}
\newcommand{\lfgtimg}[1]{\lfscene/gt/\lfscenenum_#1}
\newcommand{\lfimg}[3]{\lfscene/#1/\lfscenenum_#2_#3}

\newcommand{\gtfig}[1]{\includegraphics[width=\lfgtw,valign=c]{#1}}
\newcommand{\mfig}[1]{\includegraphics[width=\lfmw,valign=c]{#1}}
\newcommand{\gthead}[1]{\makebox[\lfgtw][c]{#1}}
\newcommand{\mhead}[1]{\makebox[\lfmw][c]{#1}}
\newcommand{\rowname}[1]{\raisebox{-.45\height}{\rotatebox{90}{\fontsize{4}{5}\selectfont #1}}}
\newcommand{\fourrow}[4]{\begin{tabular*}{\lfblockw}{@{\extracolsep{\fill}}cccc@{}}#1&#2&#3&#4\end{tabular*}}
\newcommand{\threerow}[3]{\begin{tabular*}{\lfblockw}{@{\extracolsep{\fill}}ccc@{}}#1&#2&#3\end{tabular*}}
\newcommand{\gtrow}{\fourrow{\gtfig{\lfgtimg{rgb}}}{\gtfig{\lfgtimg{segm}}}{\gtfig{\lfgtimg{depth}}}{\gtfig{\lfgtimg{normals}}}}
\newcommand{\methodrow}[2]{\threerow{\mfig{\lfimg{#1}{#2}{segm}}}{\mfig{\lfimg{#1}{#2}{depth}}}{\mfig{\lfimg{#1}{#2}{normals}}}}

\setlength{\tabcolsep}{0pt}
\renewcommand{\arraystretch}{0.6}

\textbf{Scene 0}\\[0.3ex]

\resizebox{\textwidth}{!}{
\begin{tabular}{@{}c@{\hspace{10pt}}c@{}}
\begin{tabular}{@{}c@{\hspace{2pt}}c@{}}
    & \fourrow{\gthead{Rgb}}{\gthead{Segm.}}{\gthead{Depth}}{\gthead{Normals}} \\
    \rowname{Oracle}               & \gtrow \\[0.3ex]
    & \threerow{\mhead{Segm.}}{\mhead{Depth}}{\mhead{Normals}} \\
    \rowname{Dino-Foresight}   & \methodrow{df}{short} \\
    \rowname{VFMF}             & \methodrow{vfmf}{short} \\
    \rowname{Latent-Foresight} & \methodrow{lf}{short} \\
    & (a) Short-Term \\
\end{tabular}
&
\begin{tabular}{@{}c@{\hspace{2pt}}c@{}}
    & \fourrow{\gthead{Rgb}}{\gthead{Segm.}}{\gthead{Depth}}{\gthead{Normals}} \\
    \rowname{Oracle}               & \gtrow \\[0.3ex]
    & \threerow{\mhead{Segm.}}{\mhead{Depth}}{\mhead{Normals}} \\
    \rowname{Dino-Foresight}   & \methodrow{df}{mid} \\
    \rowname{VFMF}             & \methodrow{vfmf}{mid} \\
    \rowname{Latent-Foresight} & \methodrow{lf}{mid} \\
    & (b) Mid-Term \\
\end{tabular}
\end{tabular}
}
    \vspace{-4pt}
    \caption{\textbf{Qualitative comparison on Scene 0.} We compare Dino-Foresight, VFMF, and Latent-Foresight (Ours) against ground truth (GT) for semantic segmentation, depth, and surface normal prediction at short-term (a) and mid-term (b) horizons. Latent-Foresight yields predictions that more closely match the oracle, particularly around dynamic foreground objects.}
    \label{fig:qual-0}
\end{figure}

\begin{figure}[ht]
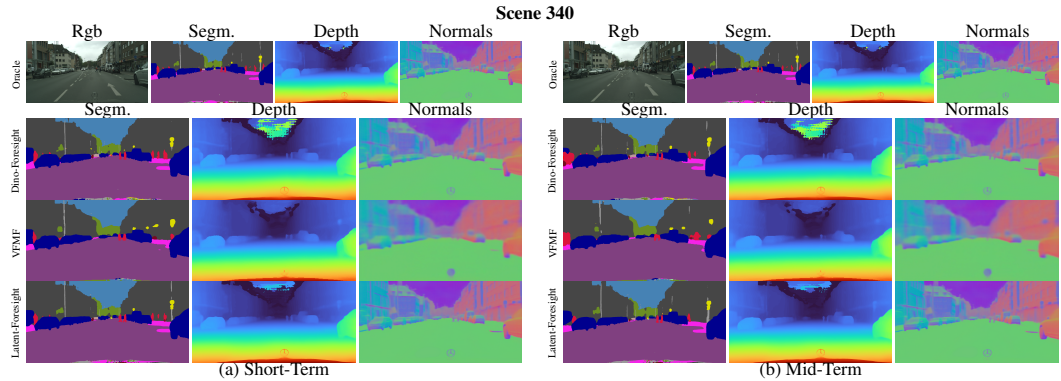

    \centering
    \scriptsize
\centering

\newcommand{\lfscene}{figs/scene_340}
\newcommand{\lfscenenum}{340}
\newcommand{\lfgtimg}[1]{\lfscene/gt/\lfscenenum_#1}
\newcommand{\lfimg}[3]{\lfscene/#1/\lfscenenum_#2_#3}

\newcommand{\gtfig}[1]{\includegraphics[width=\lfgtw,valign=c]{#1}}
\newcommand{\mfig}[1]{\includegraphics[width=\lfmw,valign=c]{#1}}
\newcommand{\gthead}[1]{\makebox[\lfgtw][c]{#1}}
\newcommand{\mhead}[1]{\makebox[\lfmw][c]{#1}}
\newcommand{\rowname}[1]{\raisebox{-.45\height}{\rotatebox{90}{\fontsize{4}{5}\selectfont #1}}}
\newcommand{\fourrow}[4]{\begin{tabular*}{\lfblockw}{@{\extracolsep{\fill}}cccc@{}}#1&#2&#3&#4\end{tabular*}}
\newcommand{\threerow}[3]{\begin{tabular*}{\lfblockw}{@{\extracolsep{\fill}}ccc@{}}#1&#2&#3\end{tabular*}}
\newcommand{\gtrow}{\fourrow{\gtfig{\lfgtimg{rgb}}}{\gtfig{\lfgtimg{segm}}}{\gtfig{\lfgtimg{depth}}}{\gtfig{\lfgtimg{normals}}}}
\newcommand{\methodrow}[2]{\threerow{\mfig{\lfimg{#1}{#2}{segm}}}{\mfig{\lfimg{#1}{#2}{depth}}}{\mfig{\lfimg{#1}{#2}{normals}}}}

\setlength{\tabcolsep}{0pt}
\renewcommand{\arraystretch}{0.6}

\textbf{Scene 340}\\[0.3ex]

\resizebox{\textwidth}{!}{
\begin{tabular}{@{}c@{\hspace{10pt}}c@{}}
\begin{tabular}{@{}c@{\hspace{2pt}}c@{}}
    & \fourrow{\gthead{Rgb}}{\gthead{Segm.}}{\gthead{Depth}}{\gthead{Normals}} \\
    \rowname{Oracle}               & \gtrow \\[0.3ex]
    & \threerow{\mhead{Segm.}}{\mhead{Depth}}{\mhead{Normals}} \\
    \rowname{Dino-Foresight}   & \methodrow{df}{short} \\
    \rowname{VFMF}             & \methodrow{vfmf}{short} \\
    \rowname{Latent-Foresight} & \methodrow{lf}{short} \\
    & (a) Short-Term \\
\end{tabular}
&
\begin{tabular}{@{}c@{\hspace{2pt}}c@{}}
    & \fourrow{\gthead{Rgb}}{\gthead{Segm.}}{\gthead{Depth}}{\gthead{Normals}} \\
    \rowname{Oracle}               & \gtrow \\[0.3ex]
    & \threerow{\mhead{Segm.}}{\mhead{Depth}}{\mhead{Normals}} \\
    \rowname{Dino-Foresight}   & \methodrow{df}{mid} \\
    \rowname{VFMF}             & \methodrow{vfmf}{mid} \\
    \rowname{Latent-Foresight} & \methodrow{lf}{mid} \\
    & (b) Mid-Term \\
\end{tabular}
\end{tabular}
}
    \vspace{-4pt}
    \caption{\textbf{Qualitative comparison on Scene 340.} We compare Dino-Foresight, VFMF, and Latent-Foresight (Ours) against ground truth (GT) for semantic segmentation, depth, and surface normal prediction at short-term (a) and mid-term (b) horizons. Latent-Foresight produces sharper and more accurate predictions than the two-stage baselines across all three tasks.}
    \label{fig:qual-340}
\end{figure}

We provide additional qualitative comparisons between Latent-Foresight and the two-stage baselines, Dino-Foresight and VFMF, on Cityscapes validation scenes. Figure~\ref{fig:qual-0} and Figure~\ref{fig:qual-340} show predictions for two representative scenes, at both short-term (a) and mid-term (b) horizons, across semantic segmentation, depth, and surface normal prediction. Across both scenes, Latent-Foresight consistently produces better predictions than the two-stage baselines: for segmentation and depth, it more accurately predicts the position and motion of dynamic objects, while for surface normals, it yields sharper predictions. 

\subsection{Definitions of Evaluation Metrics}
\label{sec:appendix-evalmetrics}

\paragraph{Movable Object Classes.}
For semantic segmentation, movable object (MO) classes comprise person, rider, car, truck, bus, train, motorcycle, and bicycle.

\paragraph{Prediction Horizons.}
On \textbf{Cityscapes}, we report short-term (3 frames, ${\sim}0.18$s) and mid-term (9 frames, ${\sim}0.54$s) prediction. On \textbf{nuScenes}, which captures more static scenes with slower dynamics, we report mid-term (9 frames, ${\sim}0.75$s), long-term (18 frames, ${\sim}1.5$s), and longer-term (27 frames, ${\sim}2.25$s) prediction. On \textbf{Kubric}, following the VFMF~\citep{boduljak2025vfm} evaluation protocol, we report 1-step (${\sim}0.17$s) and 8-step (${\sim}1.33$s) prediction.

For \textbf{depth prediction}, we report mean Absolute Relative Error $\text{AbsRel} = \frac{1}{M} \sum_{i=1}^{M} \frac{|a_i - b_i|}{b_i}$, where $a_i$ and $b_i$ are the predicted and ground truth depths at pixel $i$, and threshold accuracy $\delta_1$, the percentage of pixels satisfying $\max\!\left(\frac{a_i}{b_i}, \frac{b_i}{a_i}\right) < 1.25$. For \textbf{surface normal prediction}, we report the mean angular error $\text{m} = \frac{1}{N} \sum_{i=1}^{N} \cos^{-1}\!\left( \frac{\mathbf{n}_i \cdot \tilde{\mathbf{n}}_i}{\|\mathbf{n}_i\| \|\tilde{\mathbf{n}}_i\|} \right)$, where $\mathbf{n}_i$ and $\tilde{\mathbf{n}}_i$ are the predicted and ground truth normals, and the percentage of pixels with angular error below $11.25^\circ$, computed as $\frac{1}{N} \sum_{i=1}^{N} \mathbb{I}(\theta_i < 11.25^\circ)$.

\section{Limitations and Future Work}
\label{sec:limitations}

While our end-to-end formulation focuses on the joint training of tokenizer and flow-based predictor, the autoencoder architecture itself is kept lightweight. Future work could explore more expressive tokenizer architectures, as well as strategies for reducing the number of spatial tokens, for instance through learned spatial pooling or cross-attention-based compression, which would reduce the sequence length seen by the predictor and enable more computationally efficient generative modeling at higher resolutions.

In addition, although the current framework focuses on next-frame prediction, 
the model design naturally lends itself to broader temporal modeling strategies.
A natural extension would be to jointly model multiple future frames in a single forward pass, for instance through multi-step flow matching or latent sequence generation.

\section{Broader Impact}
\label{sec:broader_impact}
Our work advances efficient and scalable semantic future prediction by jointly learning a compact latent representation and a generative predictor over semantically rich VFM features. The resulting framework integrates flexibly with diverse scene understanding tasks without retraining, making it directly applicable to domains such as autonomous driving and robotics. By replacing two-stage pipelines with a single end-to-end training paradigm, our approach also reduces the complexity and  overhead required to develop and deploy world models. We do not foresee direct misuse risks in our work. However, as our method builds upon pretrained VFMs, any biases embedded in those models may propagate into the predicted semantic representations, potentially affecting downstream decisions in deployment scenarios. We encourage practitioners to carefully evaluate such biases before deploying systems that rely on future predictions for real-world decision-making.

\end{document}